\documentclass[sigconf]{aamas} 
\usepackage{balance} % for balancing columns on the final page
\usepackage{algorithmic}
\usepackage[linesnumbered,ruled,vlined]{algorithm2e}
\usepackage{amsmath,amsthm}
\DeclareMathOperator*{\argmin}{argmin}
\DeclareMathOperator*{\argmax}{argmax}
\newtheorem{definition}{Definition}
\newtheorem{proposition}{Proposition}
\usepackage{float}
\usepackage{newfloat}
\usepackage{listings}
\usepackage{color}
\usepackage{subcaption}
\usepackage{multirow}
\setcopyright{ifaamas}
\acmConference[AAMAS '24]{Proc.\@ of the 23rd International Conference
on Autonomous Agents and Multiagent Systems (AAMAS 2024)}{May 6 -- 10, 2024}
{Auckland, New Zealand}{N.~Alechina, V.~Dignum, M.~Dastani, J.S.~Sichman (eds.)}
\copyrightyear{2024}
\acmYear{2024}
\acmDOI{}
\acmPrice{}
\acmISBN{}

\acmSubmissionID{1232}

\title[AAMAS-2024]{Regret-based Defense in Adversarial Reinforcement Learning}
\subtitle{AAMAS 2024 --- AAAI Track}

\author{Roman Belaire}
\affiliation{
  \institution{Singapore Management University}
  \country{Singapore}}
\email{rbelaire.2021@phdcs.smu.edu.sg}

\author{Pradeep Varakantham}
\affiliation{
  \institution{Singapore Management University}
  \country{Singapore}}
\email{pradeepv@smu.edu.sg}

\author{Thanh Nguyen}
\affiliation{
  \institution{University of Oregon}
  \city{Eugene, OR}
  \country{United States}}
\email{thanhhng@cs.uoregon.edu}

\author{David Lo}
\affiliation{
  \institution{Singapore Management University}
  \country{Singapore}}
\email{davidlo@smu.edu.sg}

\makeatletter
\gdef\@copyrightpermission{
	\begin{minipage}{0.7\columnwidth}
		\href{https://creativecommons.org/licenses/by/4.0/}{This work is licensed under a Creative Commons Attribution International 4.0 License.}
	\end{minipage}
	\vspace{5pt}
}
\makeatother
\begin{abstract}
Deep Reinforcement Learning (DRL) policies are vulnerable to adversarial noise in observations, which can have disastrous consequences in safety-critical environments. For instance, a self-driving car receiving adversarially perturbed sensory observations about traffic signs (e.g., a stop sign physically altered to be perceived as a speed limit sign) can be fatal. 
Leading existing approaches for making RL algorithms robust to an observation-perturbing adversary have focused on (a) regularization approaches that make expected value objectives robust by adding adversarial loss terms; or (b) employing ``maximin'' (i.e., maximizing the minimum value) notions of robustness. While regularization approaches are adept at reducing the probability of successful attacks, their performance drops significantly when an attack is successful. On the other hand, maximin objectives, while robust, can be extremely conservative. To this end, we focus on optimizing a well-studied robustness objective, namely regret. To ensure the solutions provided are not too conservative, we optimize an approximation of regret using three different methods. We demonstrate that our methods outperform existing best approaches for adversarial RL problems across a variety of standard benchmarks from literature.
\end{abstract}


\newcommand{\BibTeX}{\rm B\kern-.05em{\sc i\kern-.025em b}\kern-.08em\TeX}

\begin{document}

%%% The following commands remove the headers in your paper. For final 
%%% papers, these will be inserted during the pagination process.

\pagestyle{fancy}
\fancyhead{}

%%% The next command prints the information defined in the preamble.

\maketitle 
%%%%%%%%%%%%%%%%%%%%%%%%%%%%%%%%%%%%%%%%%%%%%%%%%%%%%%%%%%%%%%%%%%%%%%%%

\section{Introduction}

%Deep neural networks (DNNs) as function approximation tools have allowed super-human performance across challenging domains such as games \cite{berner2019dota,mnih2013}, robotics \cite{rajeswaran2017learning,zhao2020sim}, and large-scale control \cite{huang2019adaptive,rasheed2020deep, merge}. 

Harnessing the power of Deep Neural Networks in DRL \cite{mnih2013} allows RL models to achieve outstanding results on complex and even safety-critical tasks, such as self-driving \cite{spielberg2019carla,kiran2021deep}.
However, DNN performance is known to be vulnerable to attacks on input, consequently impacting DRL models which rely on them  \cite{gleaveatk,Sun_Zhang_Xie_Ma_Zheng_Chen_Liu_2020,pattanaik2017robust}. 
In such perturbations, adversaries only alter the observations received by an RL agent and not the underlying state or the dynamics (transition function) of the environment. Even with limited perturbations, high-risk tasks such as self-driving yield opportunities for significant harm to property and loss of life. One such example is presented by \cite{shapeshifter}, in which a stop sign is both digitally and physically altered to attack an object recognition model. 

Although the effectiveness of known adversarial examples can be mitigated by adversarial training (supervised training against adversarial examples) \cite{gleaveatk,goodfellow,pattanaik2017robust,sun2023strongest}, this does not guarantee the ability to generalize to unseen adversaries. Additionally, it has been shown that naive adversarial training in RL leads to unstable training and lowered agent performance, if effective at all \cite{zhangsadqn}. 
Thus, we need algorithms that are not tailored to specific adversarial perturbations but are inherently robust. Rather than develop a policy that is value-optimal for as many known examples as possible, we want to determine what behavior and states involve risk and reduce them across the horizon. %Envision this as walking along the inside of a pathway that is at the edge of a cliff, rather than along the edge. In doing so, we minimize the exposure to effective adversarial attacks. 
To achieve this, maximin methods operate to maximize the minimum reward of a policy \cite{everett:carrl, liang2022efficient}, which can be robust but often trades unperturbed solution quality to improve the lower bound. 
Regularization methods construct adversarial loss terms to ensure actions remain unchanged across similar inputs \cite{oikarinen2021robust, zhangsadqn, liang2022efficient}, reducing the probability of a successful adversarial attack. However, as we empirically show later, they remain vulnerable when attacks are successful. 

To these ends, we provide a regret-based adversarial defense approach that aims to reduce the impact of a successful attack without being overly conservative. 

\noindent \textbf{Through our contributions, we:}
\begin{itemize}
    \item Formally define regret in settings where an observation-perturbing adversary is present. Broadly, regret is the difference of value achieved in the absence versus in the presence of an observation-perturbing adversary. 
    \item Derive an approximation of regret, named Cumulative Contradictory Expected Regret (CCER), which is amenable to scalable optimization due to satisfying the optimal sub-structure property. We provide a value iteration approach to minimize CCER, named RAD-DRN (Regret-based Adversarial Defense with Deep Regret Networks). Using CCER, our approach can balance robustness and nominal solution quality.
    \item Provide a policy gradient approach to minimize CCER, named RAD-PPO. We derive the policy gradient w.r.t the regret measure and utilize it in the PPO framework.  
    \item Provide a Cognitive Hierarchy Theory based approach named RAD-CHT that generates potential adversarial policies and computes a regret-based robust response, to demonstrate an application of CCER in adversary-reactive frameworks.
    \item Finally, we provide detailed experimental results on multiple benchmark problems (MuJoCo, Atari, and Highway) to demonstrate the utility of our approaches against several leading approaches in adversarial RL. Like previous approaches, we demonstrate the performance of our approach against strong greedy attacks (e.g., PGD). {Unlike existing work, we also show effectiveness against multi-step attack strategies that are directly computed against victim policy.}
\end{itemize}

%%%%%%%%%%%%%%%%%%%%%%%%%%%%%%%%%%%%%%%%%%%%%%%%%%%%%%%%%%%%%%%%%%%%%%%%

\section{Related Work}
\noindent\textbf{Adversarial Attacks in RL. }
Deep RL has shown to be vulnerable to attacks on its input, whether from methods with storied success against DNNs such as an FGSM attack \cite{huang,goodfellow}, tailored attacks against the value function \cite{kos,Sun_Zhang_Xie_Ma_Zheng_Chen_Liu_2020}, or adversarial behavior learned by an opposing policy \cite{gleaveatk, everett:carrl, oikarinen2021robust, zhangsadqn}. We compile attacks on RL loosely into two groups of learned adversarial policies: observation poisonings~\cite{gleaveatk,Sun_Zhang_Xie_Ma_Zheng_Chen_Liu_2020, lin:tactics2017} and direct ego-state disruptions~\cite{pinto2017robust, rajeswaran2017epopt}. Each category has white-box counterparts that leverage the victim's network gradients to generate attacks \cite{goodfellow, oikarinen2021robust, huang, everett:carrl}. While previous methods focus on robustness against one or the other, we demonstrate that the proposed methods are comparably robust to both categories of attacks. 
% In a realistic setting, an adversary may not be able to disrupt the victim's state directly (so as to remain undetected) and must provide adversarial examples as observations instead. 
%Critically, our robust methods are trained \emph{without} the presence of an adversary, in a single-agent environment.

\noindent\textbf{Adversarial Training. }
% A popular approach to robustness in RL and other deep learning domains is adversarial training, where adversarial examples are found or generated and integrated into the set of training inputs \cite{shafahi2019adversarial,ganin2016domain,wong2020fast,madry2018:adv,andriushchenko2020understanding,shafahi2020universal}. 
In this area, adversarial examples are found or generated and integrated into the set of training inputs \cite{shafahi2019adversarial, ganin2016domain, wong2020fast, madry2018:adv, andriushchenko2020understanding, shafahi2020universal}.
For a comprehensive review, we refer readers to \cite{bai2021recent}. In RL, research efforts have demonstrated the viability of training RL agents against adversarial examples \cite{gleaveatk, bai2019model, pinto2017robust, tan2020robustifying, kamalaruban2020robust}. Naively training RL agents against known adversaries is a sufficient defense against known attacks; however, new or more general adversaries remain effective \cite{gleaveatk, kang2019transfer} and therefore we focus on proactively robust defense methods instead of reactive (react to known adversaries) defense methods. %PGD-based adversarial training \cite{kamalaruban2020robust,tan2020robustifying} is more effective in this regard, but critically cannot provide guarantees on robustness \cite{kang2019transfer}. 

\noindent\textbf{Robustness through Regularization.}
%Other approaches to achieving adversarial robustness seek policies trained to optimize theoretically-supported robustness measures. 
Regularization approaches ~\cite{zhangsadqn, oikarinen2021robust, everett:carrl} take vanilla value-optimized policies and robustify them to minimize the loss due to adversarial perturbations. These approaches utilize certifiable robustness bounds computed for neural networks when evaluating adversarial loss and ensure the probability of success of an attack is reduced using these lower bounds.   %sacrifice optimal unperturbed performance to fortify the lower bound. %Regularization methods \cite{oikarinen2021robust,zhangsadqn} enforce value-optimal policies to have similar outputs given similar inputs; however, 
{\em Despite lowering the likelihood of a successful attack, an attack that does break through will still be effective (as shown in Table \ref{tab:atk success})}. Note that for RADIAL (a regularization approach), even though the success percentage of attacks is the least, it has the largest drop in performance. 
%A class of defenses using certifiable robustness \cite{zhangsadqn,oikarinen2021robust,everett:carrl} computes lower bounds of policy networks to reduce adversarial exposure. We compare to RADIAL \cite{oikarinen2021robust}, and CARRL \cite{everett:carrl} as baselines. Lower bound methods can be quite robust but oftentimes sacrifice optimal unperturbed performance to fortify the lower bound. Regularization methods \cite{oikarinen2021robust,zhangsadqn} enforce value-optimal policies to have similar outputs given similar inputs; however, despite lowering the likelihood of a successful perturbation, an attack that does break through will still be effective (as demonstrated in Table \ref{tab:atk success}). Most similar to our work is \cite{liang2022efficient} which, building on the above methods, directly learns and optimizes the worst-case return using the Bellman equation and constructs a worst-attack regularization term. Despite being state-of-the-art, we find that it is not a holistic solution for the limitations of maximin methods.

\begin{table}[t!]
    \centering
    \begin{tabular}{ |p{1.8cm}|p{1.8cm}|p{1.8cm}|p{1.3cm}| }\hline
    Algo. & Num. Atks & \% Success & Score\\\hline
    RADIAL      &  6    &   5\%    &   4.16\\
    WocaR       & 6     &   12\%    &   6.63\\
    RAD-DRN     &6      &   40\%     &   9.9\\
    RAD-PPO     &7      &   10\%     &   18\\
    RAD-CHT     &8      &   60\%     &   20.1\\\hline
    \end{tabular}
    \caption{The impact and frequency of successful attacks on an example set of the Strategically Timed Attack \cite{lin:tactics2017} on \textit{highway-fast-v0}. The Num. Atks column shows the number of attempted attacks; the \% Success column indicates the rate at which an attempted attack changed the selected action. When unperturbed, all methods reach approximately 22 points.}
    \label{tab:atk success}
\end{table}

\noindent\textbf{Regret Optimization in MDPs.}
Measuring and optimizing a regret value to improve the robustness has been studied previously in uncertain Markov Decision Processes (MDPs)\cite{ahmed2013regret, Rigter_Lacerda_Hawes_2021}. In RL, \cite{jin2018regret} established Advantage-Like Regret Minimization (ARM) as a policy gradient solution for agents robust to partially observable environments. While the properties and optimization of regret are in general well-studied, {\textit{current applications in RL focus on utilizing regret as a robustness tool against natural environment variance; to the best of our knowledge, this is the first application of regret to defend against strategic adversarial perturbations}}.

{ \noindent\textbf{Adversary-Agnostic Approaches}
Unlike adversary-specific robustness training, the methods we term as "adversary-agnostic" do not interact with a perturbed MDP during training. While the various forms of adversarial retraining do have merit, they often take longer to train (needing to train both victim and adversary policies). PA-ATLA-PPO \cite{sun2023strongest}, a SoTA adversarial retraining technique, reports needing 2 million training frames for MuJoCo-Halfcheetah. For comparison, both our proposed RAD-PPO and WocaR-PPO \cite{liang2022efficient}, another SoTA adversary-agnostic method, require less than 40\% of the training frames. Furthermore, adversary-specific methods may have subpar performance against novel adversaries, in addition to well-known drawbacks such as catastrophic forgetting. \cite{zhangsadqn} demonstrates how naive adversarial retraining is in general a poor solution; we further demonstrate in Table \ref{tab:agnostic vs specific} that even advanced retraining frameworks are not as generally robust.

\begin{table}[]
    \centering
    \begin{tabular}{|c|c|c|c|}
    \hline
    Environment & Algorithm & Unperturbed & Random Pert. \\\hline
         \multirow{3}{*}{Hopper} & PA-ATLA-PPO & 3449 & 1564 \\
                                  &  WocaR-PPO & 3136 & 3242\\
                                  &  RAD-PPO     & 3473 & 3415 \\\hline
         \multirow{3}{*}{HalfCheetah} & PA-ATLA-PPO & 6289 & 3414 \\
                                  &  WocaR-PPO & 3993 & 4128\\
                                  &  RAD-PPO     & 4426 & 4387 \\\hline
    \end{tabular}
    \caption{Adversary-specific robustness methods (forms of adversarial retraining) are robust only to the adversarial behavior they specifically train against. PA-ATLA-PPO is the SoTA adversarial retraining method, and WocaR-PPO is the SoTA adversary-agnostic method. RAD-PPO is our best-performing adversary-agnostic approach. Note that the adversary-specific models perform poorly against even random adversarial perturbations, which are weaker than other attacks. Attacks use perturbation radius $\epsilon=0.15$.}
    \label{tab:agnostic vs specific}
\end{table}

}
%%%%%%%%%%%%%%
%%%%%%%%%%%%%%
\section{RL with Adversarial Observations}
In problems of interest, we have an RL agent whose state/observation and action space are $S$ and $A$ respectively. There is an underlying transition function $T:S \times A \times S \rightarrow [0,1]$ and reward model $R:S \times A \rightarrow \mathbb{R}$ which are not known {\em a priori} to the RL agent.  The behavior of the agent is governed by a policy $\pi: S \rightarrow A$ that maps states to actions.
The typical objective for an RL agent is to learn a policy ${\pi}^*$ that maximizes its expected value without knowing the underlying transition and reward model {\em a priori}. This can be formulated as the following optimization problem:
\begin{align*}
    & {\pi}^* \in \arg\max\nolimits_{{\pi}} V^\pi(s_0)
\end{align*}
where $V^\pi(s_0)$ is the expected accumulated value the RL agent obtains starting from state $s_0$ for executing policy $\pi$.
The policy computed is not robust in the presence of an adversary who can strategically alter the observation received by the agent at any time step. This is because the agent could be executing the wrong action for the underlying state.\footnote{It should be noted that an adversary is assumed to alter only the agent observations and not the model dynamics or the real states.} 

For the rest of this paper, we will use $z_t$ to represent the observed (possibly perturbed) state and $s_t$ to refer to the true underlying state at step $t$.
Formally, on taking action $a_t$ in state $s_t$, the environment transitions to a new state $s_{t+1}$. However, the adversary can alter the observation received by the RL agent to another state $z_{t+1}$ instead of $s_{t+1}$ to reduce the expected value of the RL agent, where $z_{t+1} \in N(s_{t+1})$ --- a set of neighbors of $s_{t+1}$, defined as follows: 
$$N(s_t) = \{z_t: \Vert {z}_t-s_t \Vert_{\infty} \leq \epsilon \} $$
Often, it is not possible to discern if the observation is perturbed or not. For example, given a highway speed limit sign, a perturbation showing a different speed limit may be undetectable for an agent and hence would be within the neighborhood. However, this is within a reasonable constraint; perturbing the speed limit sign to be a stop sign may be easy to heuristically distinguish and hence would not be part of the neighborhood.

We now can represent the \emph{observation-altering adversarial policy} as $\mu:S \rightarrow S$. The notion of neighborhoods translates to adversarial policies $\mu$ such that $\mu(s) \in N(s)$. When the adversarial policy is deterministic and bijective, for a given observed state ${z}$, we will employ $\mu^{-1}({z})$ to retrieve the underlying unperturbed state $s\in N(z)$. 
In the following, we focus on computing policies that are inherently robust, even without knowing the adversary policy.

\section{Regret-based Adversarial Defense (RAD)}
Our approach to computing inherently robust policies is based on minimizing the maximum \emph{regret} the agent receives for taking the wrong action assuming there was a perturbation in states. We first introduce the notions of regret and max regret for the RL agent to play a certain policy $\pi$. The regret-based robust policy is then formulated accordingly.\footnote{Our regret and minimax regret policy are defined w.r.t deterministic policies $\pi$ and $\mu$, for the sake of representation. Extending these definitions for stochastic policies is straightforward.}

\begin{definition}[Regret]
Given an adversary policy $\mu$, the regret at each observed state $z_t$ for the RL agent to play a policy $\pi$ is defined as follows:
\begin{align}
    &\delta^{\pi,\mu}(z_t) = V^{\pi}(z_t) - V^{\pi, \mu}(z_t) \label{eq:reg2}
\end{align}
\end{definition} 

Intuitively, it is the difference between the expected value $V^\pi(\cdot)$ (assuming no adversary) and the value $V^{\pi, \mu}(\cdot)$ (assuming states are being perturbed by the adversary policy $\mu$) while the RL agent takes actions according to a policy $\pi$. 
In particular, the value for an agent with a policy $\pi$ in the absence of an adversary, $V^\pi(\cdot)$, is given by:
\begin{align}
 V^{\pi}(z_t) &= R(z_t, \pi(z_t)) + \gamma \mathbb{E}_{z_{t+1}} \big[V^{\pi}(z_{t+1})\big] \label{eq:reg3}
 \end{align}
where $z_{t+1} \sim T(\cdot|z_t, \pi(z_t))$. Note that when the adversary is not present, the observed states $z_t = s_t$ and $z_{t+1}=s_{t+1}$.

{On the other hand, the $V^{\pi,\mu}(.)$ function for an agent with policy $\pi$ in the presence of such an adversary $\mu$ is given by:}
%  \begin{align}
% V^{\pi, \mu}(z_{t}) &= R(z_{t},\pi(\mu(z_t))) + \gamma \mathbb{E}_{z_{t+1}} \Big[V^{\pi,\mu}({s}_{t+1})\Big] \label{eq:reg4}
% \end{align}
 \begin{align}
V^{\pi, \mu}(z_{t}) &\!=\! R(s_t,\pi(z_t)) \!+\! \gamma \mathbb{E}_{z_{t+1}} \big[V^{\pi,\mu}({z}_{t+1})\big] \label{eq:reg4}
\end{align}
where $s_t\!\! =\!\! \mu^{-1}\!(z_{t})$, $z_{t+1}\!\!=\!\!\mu(s_{t+1})$ with $s_{t+1}\!\!\sim\!\! T(\cdot| s_t, \pi(z_t))$.
%where $z_{t+1} = \mu({s}_{t+1})$ and $s_{t+1}\sim P(\cdot|\mu^{-1}(z_t), \pi(z_t))$.
 Note that our regret is defined based on observed states, as the agent will only have access to observed states (and not the true states) when making test-time decisions.
\begin{definition}[Minimax Regret Policy]
    The regret-based robust policy for the RL agent is the policy that minimizes the maximum regret at the initial observed state $z_0$ over all possible adversary policies, formulated as follows:
    \begin{align}
        &{\pi}^* \in \arg\min\nolimits_{{\pi}}\; \max\nolimits_{\mu}\; \delta^{\pi,\mu}(z_0) \label{eq:reg1}
    \end{align}
\end{definition}
Intuitively, a minimax regret policy minimizes the maximum regret by avoiding actions that have high variance across neighboring states. 
% \begin{align}
%     &\max_\mu \; reg^{\pi, \mu}(z_0) =V^{\pi}(z_0) -\min_\mu  V^{\pi, \mu}(z_0)\nonumber\\
% \intertext{where}
% &V^{\pi,\mu}(z_t) = R(z_t, \pi(\mu(z_t))) + \gamma \mathbb{E}_{{s}_{t+1}} {V}^{\pi,\mu}({s}_{t+1})\label{eq:reg6}
%      %&\text{with } s_{t+1} \sim T(\cdot|\mu^{-1}(z_t), \pi(z_t)),z_{t+1}=\mu(s_{t+1})\nonumber
% \end{align}
Unfortunately, there are multiple issues with optimizing minimax regret. First, optimizing regret requires iterating over different adversary policies, which can be infinite depending on the state and action spaces. If we assume specific types of adversaries, robustness may not extrapolate to other adversary policies. Second, the minimax regret expression does not exhibit optimal substructure property, thus rendering value iteration approaches (e.g., Q-learning, DQN) theoretically invalid. Third, deriving policy gradients w.r.t. regret is computationally challenging, requiring combinatorial perturbed trajectory simulations. As a result, employing policy gradient approaches to optimize regret is also infeasible. 

We address the above issues by using an approximate notion of regret referred to as Cumulative Contradictory Expected Regret (CCER) and then proposing two types of approaches (adversary agnostic\footnote{ Agnostic methods do not receive perturbations while training.} and adversary dependent):
\begin{itemize}
\item Adversary agnostic approach for optimizing approximate regret: CCER has multiple useful properties: (i) It satisfies the optimal substructure property, thereby allowing for usage of a DQN type approach; and (ii) It is possible to compute a policy gradient with regards to CCER, thereby allowing for a policy gradient approach. 
\item Adversary dependent approach for optimizing approximate regret: We utilize an iterative best-response approach based on Cognitive Hierarchical Theory (CHT) to ensure defense against a distribution of ``good'' adversarial policies. This approach is referred to as RAD-CHT.
\end{itemize}

% \begin{align}
%     &\max_\mu \; reg^{\pi, \mu}(z_0) \approx  V^{\pi}(z_0) - \check{V}^{\pi}(z_0) \label{eq:reg5}\\
% \intertext{where }
%  &\check{V}^{\pi}(z_t) = \min_{s_t\in N(z_t)} R(s_t, \pi(z_t)) + \gamma \mathbb{E}_{{s}_{t+1} } \check{V}^{\pi}({s}_{t+1})\label{eq:reg7}\\
%  &\text{with }{s}_{t+1} \sim T(\cdot|z_t, \pi(z_t))\nonumber
% \end{align}

%Since we want to find a robust policy for \emph{any} adversary, we compute the maximum regret (in Equation~\ref{eq:reg1}) by finding an upper bound on regret (aka. maximum regret) over an assumed set of all possible adversarial perturbations. 
% allowing us to make this computation when $\mu$ or $\mu^{-1}$ is not known. 
%There are two key conditions assumed when constructing this upper bound:\\
%\noindent (a) The received state is always perturbed, at current and for future steps (upper bound on frequency). \\
%\noindent (b) The neighborhood $N({{s}_t})$ encompasses the maximum perturbation an adversary will provide at $z_t$ (upper bound on magnitude). 
\subsection{Regret Approximation: CCER}
% {\color{blue} Thanh: Just for clarification, we are looking at the test-time attacks, right? Meaning we train our model in an unperturbed environment and then test our trained model in a perturbed environment? }
% { Please detail the definition of CCER; How to use CCER for actor-critic method?}
We introduce a new notion of approximate regret, referred to as CCER. 
% This definition does not require access to $\mu$.
CCER accumulates contradictory regret at each epoch; we refer to this regret as contradictory as the observed state's optimal action may contradict the true state's action. The key intuition is that we accumulate the regret (maximum difference in reward) in each time step regardless of whether the state is perturbed or not perturbed. 

For the underlying transition function, $T$, and reward model, $R$, CCER w.r.t a policy $\pi$ is defined as follows: 
\begin{align}\label{eq:ccer}
\delta_{\texttt{CCER}}^\pi(z_t) &= R(z_t,\pi(z_t)) - \min\nolimits_{s_t \in N({z_t})} R(s_t, \pi(z_t)) \nonumber\\
&\hspace{0.4in} + \gamma \mathbb{E}_{z_{t+1}} \Big[\delta_{\texttt{CCER}}^\pi(z_{t+1})\Big] 
\end{align}
where $z_{t+1}\sim T(\cdot|z_t, \pi(z_t))$.
Our goal is to compute a policy that minimizes CCER, i.e.,
$$\pi^{\bot} \in \arg \min\nolimits_{\pi} \;\delta_{\texttt{CCER}}^\pi(z_0)$$
This is a useful objective, as CCER accumulates the myopic regrets at each time step and we minimize this overall accumulation. Importantly, CCER has the optimal substructure property, i.e., the optimal solution for a sub-problem (from step $t$ to horizon $H$) is also part of the optimal solution for the overall problem (from step $0$ to horizon $H$). %This allows for Q-learning type learning algorithms to be employed to optimize regret when transition and reward models are not available. We can also derive policy gradients with regard to CCER more easily, so policy gradient-type approaches can also be employed. 

% {Proposition [below] seems odd. In particular, it seems odd that an optimal and a non-optimal policy behave identically for
% each state up to time t. This assumption, on which the proof is then based, is a bit strong}

\begin{proposition}[Optimal Substructure property]\label{prop.1}
At time step $t$, the CCER corresponding to a policy, $\pi$ at $z_t$, i.e., $\delta_\texttt{CCER}^{\pi}(z_t)$ is minimum if it includes the CCER minimizing policy from $t+1$, i.e.,$\pi^\bot_{[t+1,H]}$ from $t+1$. Formally, 
$$\delta_\texttt{CCER}^{\left<\pi_t, \pi^\bot_{[t+1,H]}\right>}(z_t) \leq \delta_\texttt{CCER}^{\left<\pi_t, \pi_{[t+1,H]}\right>}(z_t)$$
\end{proposition}

All proofs are in Appendix A. Similar to the \emph{state} and \emph{state-action} value function, $V(s)$ and $Q(s,a)$, we also have state regret, $\delta^{\pi}_{\texttt{CCER}}(z_t)$ and state-action regret, $\delta^{\pi}_{\texttt{CCER}}(z_t,a_t)$. Specific definitions are provided in the following sections. 

\subsection{Approach 1: RAD-DRN}

We first provide a mechanism similar to DQN~\cite{mnih2015human} to compute a robust policy that minimizes CCER irrespective of any adversary. Briefly, we accumulate the regret over time steps (rather than reward) and act on the minimum of this estimate (rather than the maximum, as a DQN would). 
Intuitively, minimizing CCER ensures the obtained policy avoids taking actions that have high variance, therefore avoiding volatile states that have high regret. In our adversarial setting, this roughly corresponds to behavior that avoids states where false observations have q-values vastly different from the underlying state. In a driving scenario, for example, this could mean giving vehicles and obstacles enough berth to account for errors in distance sensing. 
Let us denote by: 
\begin{align*}
    \delta^\pi_{\texttt{CCER}}(z,a)& = R(z, a) - \min\nolimits_{s\in N(z)}R(s, a)\\
    &\qquad\qquad+ \gamma \mathbb{E}_{z'}\left[\min\nolimits_{a'}\delta^\pi_{\texttt{CCER}}(z', a')\right]
\end{align*}
% $$Q^{ccer}(s,a) = R(s, a) - \min_{s\in N(s)}R(s, a) + \gamma \mathbb{E}_{s'}\left[\min_{a'}Q^{ccer}(s', a')\right]$$
as the \emph{q-value} associated with CCER in our setting. We provide the pseudocode of our algorithm, named RAD-DRN, employing a neural network to predict $\delta_{\texttt{CCER}}(z,a)$ in Alg.~\ref{alg:drn}. This network's parameter $w$ is trained based on minimizing the loss between the predicted and observed $\delta_{\texttt{CCER}}$ values.

\begin{algorithm}[t!]
\caption{RAD-DRN}
\label{alg:drn}
Initialize replay memory $D$ to capacity $N$;\\
 % Initialize replay memory $D$;\\
Initialize q-regret $\delta_{\texttt{CCER}}$ with random weights $w$;\\
Initialize target q-regret $\widehat{\delta}_{\texttt{CCER}}$ with weights $w^- =w$;\\
\For{$\text{episode} = 1 \to M$}{
Get initial state $z_0$;\\
\For{$t = 0 \to H$}{
With prob. $\epsilon$, select a random action $a_t$;\\
Else, select $a_t\in\argmin_{a} \delta_{\texttt{CCER}}(z_t, a,w)$;\\
Execute action $a_t$, get observed state $z_{t+1}$; \\
Observe regret $r_t \!=\! R(z_t, a_t) \!-\! \!\!\!\min\limits_{s_t\in N(z_t)}\!\!\!\!R(s_t, a)$;\\
Store transition $(z_t, a_t, z_{t+1}, r_t)\rightarrow D$;\\
Sample mini-batch $(z_i, a_i, z_{i+1}, r_i)\sim D$;\\
\For{each $(z_i, a_i, z_{i+1}, r_i)$ in mini-batch}{
$\text{Set target }y_i =
  \begin{cases}
    r_i        \text{, if episode terminates at step $i+1$} \\
    r_i + \gamma \min_{a'}\widehat{\delta}_{\texttt{CCER}}(z_{i+1},a';w^-)   \text{, otherwise }
  \end{cases}$\\
Perform a gradient descent to update $w$ based on loss: $\big[y_i - \delta_{\texttt{CCER}}\big(z_i, a_i, w\big)\big]^2$;\\
}
}
Every $K$ steps reset $w^- = w$;
}
\end{algorithm}

Given the conservative nature of regret, RAD-DRN can provide degraded nominal (unperturbed) performance, especially in settings with very few high-variance states. We propose a heuristic optimization trick to reduce unnecessary conservatism in RAD-DRN.
We consider using weighted combinations of value and CCER estimates to determine the utility of employing each policy at a given observation. Intuitively, we want to apply a value-maximizing policy in low-variance state neighborhoods and a regret-minimizing policy in high-variance neighborhoods; as such, the utility of using a regret-minimizing policy in a high-regret neighborhood will, in turn, be high. Specifically, we train a DQN to maximize the utility function: 
$$Util(z, \pi)=V^{\pi}(z)-\beta \delta^\pi_{\texttt{CCER}}(z)$$
where $\beta$ is a caution-weight constant decimal. This final step requires a minimal amount of training ($500$ episodes) and interacts directly with the non-adversarial environment. We constrain the available actions of the utility-maximizing policy to either be the value-optimal action $\argmax_a Q^\pi(z,a)$ or the CCER-optimal action $\argmin_a \delta^\pi_{\texttt{CCER}}(z,a)$, avoiding actions that are sub-optimal to both measures. These two optimal DQN and RAD-DRN policies are obtained in advance before training the above utility-maximizing policy. 

\subsection{Approach 2: RAD-PPO}
{Approach 1 provides a regret iteration approach (along the lines of value iteration in Deep Q Learning). In this section, we provide a policy gradient approach that minimizes CCER.} Proximal Policy Optimization (PPO), and any policy gradient (PG) method, can be extended with contradictory regret. The key contribution is defining/deriving the policy gradient based on the long-term CCER of a policy: 
% For ease of writing, let $\sigma_{\pi}(s,a)$ represent the soft-min reward of action $a$ at observed state $s$ in a neighborhood $N(s)$ given the policy $\pi$, which is formulated as follows:
% \begin{align*}
%     \sigma_\pi(s,a)=&\sum\limits_{\tilde{z_i}\in N(s)}\frac{e^{-R(\tilde{z_i},a)}}{\sum\limits_{\tilde{z_j}\in N(s)}e^{-R(\tilde{z_j},a)}}R(\tilde{z_i},a) 
% \end{align*}
\begin{align*}
     \delta_{\texttt{CCER}}^{\pi}(z)=&\sum\nolimits_{a}\pi(z,a)\delta_{\texttt{CCER}}^{\pi}(z,a)
\end{align*}
% \begin{definition}\label{def:PG_CCER}
%     The  differentiable form of CCER is represented as follows:
% \begin{align*}
%     Q^{CCER}_{\pi}(s)=&\sum_{a}\pi(s,a)Q^{ccer}_\pi(s,a) \\
%     Q^{ccer}_\pi(s,a)=&\mathbb{E}\left[\sum\limits_{t} \gamma^{t-1}\left(R(s^t,a^t)-\sigma(s^t,a^t)\right)\right] 
% \end{align*}
%  \text{given $ \sigma_\pi(s,a)$ is a soft-min function:}
% \begin{align*}
%     \sigma_\pi(s,a)=&\sum\limits_{\tilde{z_i}\in N(s)}\frac{e^{-R(\tilde{z_i},a)}}{\sum\limits_{\tilde{z_j}\in N(s)}e^{-R(\tilde{z_j},a)}}R(\tilde{z_i},a) 
% \end{align*}
% \end{definition}
\begin{proposition}\label{prop.2}
    The gradient of a CCER-minimizing policy $\pi_{\theta}$ parameterized by $\theta$ can be computed as follows:
    \begin{align}\label{eq:policygradient}
        \frac{\partial \delta_{\texttt{CCER}}^{\pi_\theta}(z)}{\partial\theta} &\!=\! \sum\nolimits_s \!\!P(z|\pi_\theta)\! \sum\nolimits_a\!\! \frac{\partial\pi_\theta(z,a)}{\partial\theta}\delta_{\texttt{CCER}}^{\pi_\theta} (z,a) %- uncomment this line if softmin relies on theta \pi(s,a)\frac{\partial\sigma_{\pi}(s,a)}{\partial\theta}
\end{align}
where $P(z|\pi_\theta)$ is the stationary distribution w.r.t $\pi_{\theta}$.
\end{proposition}

This can also be rewritten along similar lines as the traditional policy gradient using an expectation (instead of $P(z|\pi_\theta)$):
\begin{align*}
    & \nabla_{\theta} \delta_{\texttt{CCER}}^{\pi_\theta}(z_0) = \mathbb{E}_{\pi_\theta} \Big[\delta_{\texttt{CCER}}^{\pi_\theta}(z,a )\nabla_{\theta} \log\pi_{\theta}(a\mid z)\Big]
\end{align*}
\paragraph{CCER-based Advantage.}Furthermore, we can replace $\delta_{\texttt{CCER}}^{\pi_\theta}(z,a )$ on the right side of the gradient with a CCER advantage function defined as follows:
\begin{equation}\label{eq:Advantage}
    \begin{aligned}
        {A}^{\texttt{CCER}}(z_t,a)&=-[\delta_{\texttt{CCER}}^{\pi_\theta}(z_t,a)-\delta_{\texttt{CCER}}^{\pi_\theta}(z_t)]
    \end{aligned}
\end{equation}
The positive value of $|{A}^{\texttt{CCER}}|$ can be understood as the increase in regret associated with increasing $\pi_\theta(a|z_t)$, as the value of the selected action relative to others will be lower when a perturbation occurs. Thus, we invert the sign so that regret and ${A}^{\texttt{CCER}}$ are inversely related, i.e. selecting the action with the highest ${A}^{\texttt{CCER}}$ will minimize regret.

\subsection{Approach 3: RAD-CHT}
The previous RAD-DRN and RAD-PPO approaches are agnostic to specific adversarial policies. Here, we provide a reactive framework that identifies strong adversarial policies and computes a best-response (minimum CCER) policy against them. This approach builds on the well-known behavior model in game theory and economics referred to as Cognitive Hierarchical Theory (CHT)~\cite{camerer:cht}. In CHT, players' strategic reasoning is organized into multiple levels; players at each level assume others are playing at a lower level (i.e., are less strategic). 
 
Our approach, named RAD-CHT, is an iterative algorithm that proceeds as follows:
 \begin{itemize}
     \item Iteration 0 (Level $0$): Both the RL agent and the adversary play completely at random, with the random policies for agent and adversary given by $\pi^{(0)}$ and $\mu^{(0)}$ respectively.
     \item Iteration 1 (Level $1$): The RL agent assumes a level $0$ adversary and wants to find a new policy $\pi^{(1)}$ that minimizes the regret given $\mu^{(0)}$, formulated as follows: 
     % \begin{align*}
     %     & \pi^1 \in \argmin_{\pi} V^{\pi}(\cdot) - V^{\pi,\mu^{0}}
     % \end{align*}
    \begin{align*}
          \pi^{(1)} &= \arg\min\nolimits_{\pi} \delta^{\pi,\mu^{[(0)]}}_\texttt{CCER}(z_0) \nonumber\\
          \delta_\texttt{CCER}^{\pi,\mu^{[(0)]}}(z_t) &= R(z_t, \pi(z_t)) -  R(s_t, \pi(z_t))  \nonumber\\
        &\hspace{0.5in} +\gamma \mathbb{E}_{z_{t+1}} \left[\delta_{\texttt{CCER}}^{\pi,\mu^{[(0)]}}(z_{t+1})\right]
    \end{align*}
     where $s_t$ is the true state given the observed state is $z_t$ and the attack policy is $\mu^{(0)}$, i.e., $\mu^{(0)}(s_t) = z_t$. On the other hand, the level-$1$ adversary assumes the RL agent is at level $0$ and attempts to find a perturbation policy that minimizes the agent's expected return:
     \begin{align*}
         & \mu^{(1)} \in \arg\min\nolimits_{\mu} V^{\pi^{(0)}, \mu}(\cdot)
     \end{align*}
     \item Iteration $k > 1$ (Level $k$): The agent assumes the adversary can be at any level below $k$. Like in CHT, the adversary policy is assumed to be drawn from a Poisson distribution over the levels $0, 1, \cdots, k-1$ with a mean $k-1$. The RL agent then finds a new policy $\pi^{(k)}$ that minimizes the regret formulated as the following:
     % \begin{align*}
     %     &\pi^k \in \argmin_{\pi} V^{\pi}(\cdot) - \sum_{i = 0}^{k-1} \frac{\lambda^ie^{-\lambda}}{i!} V^{\pi, \mu^i}(\cdot)
     % \end{align*}
     {\small\begin{align*}
        \pi^{(k)} &\in \arg\min\nolimits_{\pi} \delta^{\pi,\mu^{[(k-1)]}}_\texttt{CCER}(z_0) \nonumber\\
     \delta^{\pi,\mu^{[(k-1)]}}_\texttt{CCER}(z_t) &= R(z_t, \pi(z_t)) - \!\!\sum\nolimits_{i=0}^{k-1} \!P^{(k)}\!(i) R(s_t^{(i)}\!, \pi(z_t)) \\
     &\qquad\qquad+
     \gamma \mathbb{E}_{z_{t+1}} \left[\delta_{\texttt{CCER}}^{\pi,\mu^{[(k-1)]}}(z_{t+1})\right]
    \end{align*}}
     where $P^{(k)}(i)\propto \frac{\lambda^i e^{-\lambda}}{i!}$ (aka. Poisson distribution) is the probability the adversary is at level $i$ with $0\leq i\leq k-1$. The notation $[(k-1)]$ represents the range $(0),\ldots,(k-1)$. In addition, the true state is $s_t^{(i)}$, the observed state is $z_t$ and the attack policy is $\mu^{(i)}$, i.e., $\mu^{(i)}(s_t^{(i)}) = z_t$. 
     
     Similarly, the adversary at this level assumes the RL agent is at a level below $k$, following the Poisson distribution. The adversary then optimizes the perturbation policy as follows: 
     \begin{align*}
         & \mu^{(k)} \in \argmin\nolimits_{\mu} \sum\nolimits_{i = 0}^{k-1} P^{(k)}(i) V^{\pi^{(i)}, \mu}(\cdot)
     \end{align*}
 \end{itemize}
 At each iteration, the CCER minimization problem given the adversary distribution exhibits optimal substructure. Therefore, we can adapt Approach 1 and 2 to compute the CCER-minimizing policy given the adversarial distribution. 

 On the adversary's side, we provide the policy gradient given the victim policies at previous levels. Let's denote $$V_{\mu_{\theta}}(\cdot) = \sum\nolimits_{i = 0}^{k-1} P^{(k)}(i) V^{\pi^{(i)}, \mu_{\theta}}(\cdot)$$
where $\theta$ is the parameter of the adversary policy $\mu$.
 \begin{proposition}\label{CHT.adv.gradient}
     The gradient of the objective function of the adversary $V_{\mu_{\theta}}(\cdot)$ w.r.t $\theta$ can be computed as follows:
     \begin{align*}
         & \nabla_{\theta}V_{\mu_{\theta}} \!=\! \mathbb{E}_{i\sim P^{(k)}(i), \tau\sim (\mu_{\theta}, \pi^{(i)}, T)} \Big[R(\tau) \!\sum\nolimits_t \!\!\nabla_\theta\log \mu_{\theta}(z_t|s_t)\Big]
     \end{align*}
     where $\tau = (s_0, z_0, a_0, s_1, z_1, a_1,\cdots)$ with $z_t\sim \mu_{\theta}(s_t)$ is the perturbed state created by the adversary policy $\mu_{\theta}$. 
     In addition, the reward of a trajectory $\tau$ is defined as follows:
     \begin{align*}
         & R(\tau) = \sum\nolimits_t \gamma^t R(s_t, a_t)
     \end{align*}
 \end{proposition}

 Based on Proposition~\ref{CHT.adv.gradient}, finding an optimal adversarial policy at each level can be handled by a standard policy optimization algorithm such as PPO.

\def\imgwidth{0.25}
\begin{figure*}[t]
    \begin{subfigure}[b]{\textwidth}
        \hspace{0em}\begin{subfigure}[b]{\imgwidth\textwidth}
        %\centering
            \includegraphics[width=\textwidth]{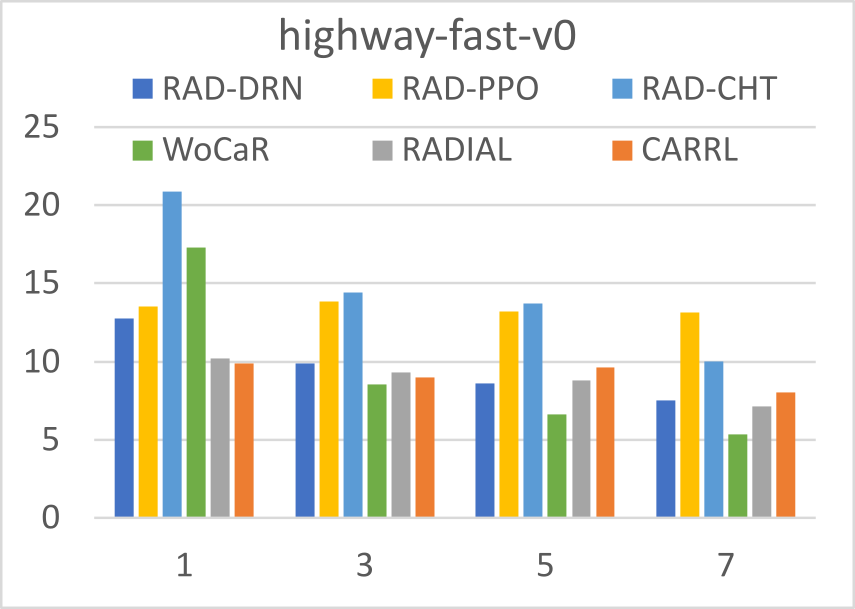}%
            \label{fig:timed:hwy}
        \end{subfigure}%
        \hspace{0em}\begin{subfigure}[b]{\imgwidth\textwidth}%
        %\centering
            \includegraphics[width=\textwidth]{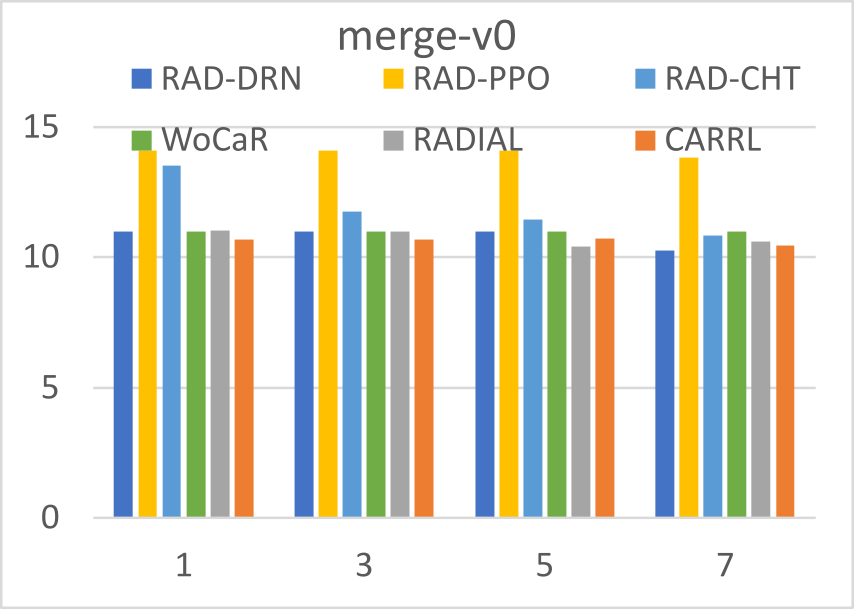}%
            \label{fig:timed:mrg}
        \end{subfigure}%
        \hspace{0em}\begin{subfigure}[b]{\imgwidth\textwidth}%
        %\centering
            \includegraphics[width=\textwidth]{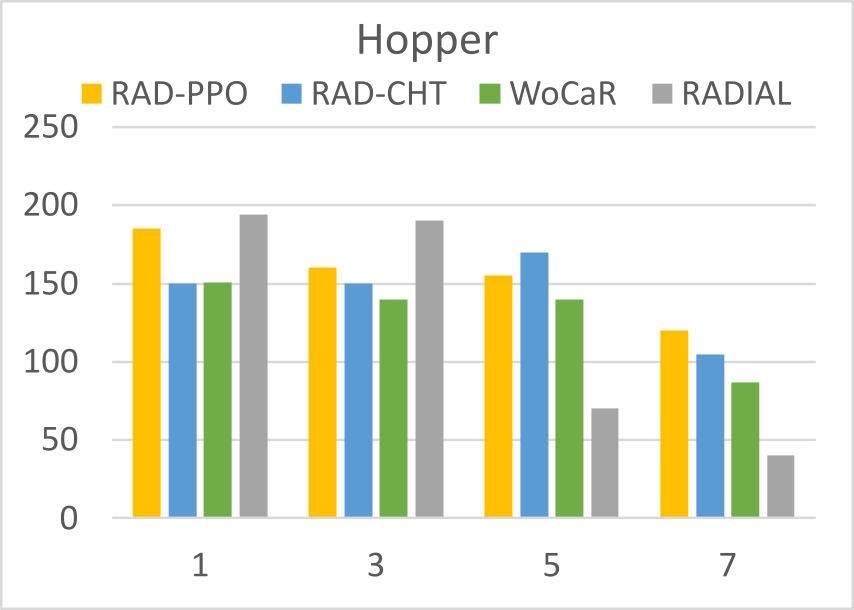}%
            \label{fig:timed:hopper}
        \end{subfigure}%
        %\hspace{-2.2em}\begin{subfigure}[b]{\imgwidth\textwidth}%
        %\centering
            %\includegraphics[width=\textwidth]{media/halfcheetah_timed.pdf}%
            %\label{fig:timed:halfcheetah}
        %\end{subfigure}%
        \begin{subfigure}[b]{\imgwidth\textwidth}%
        %\centering
            \includegraphics[width=\textwidth]{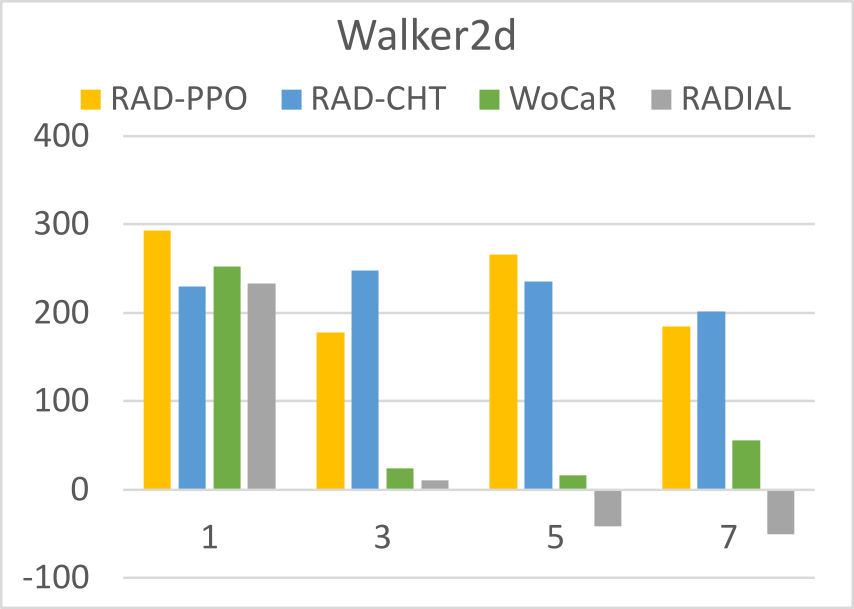}%
            \label{fig:timed:walker}
        \end{subfigure}%
        \caption{Score vs. Percentile Perturbation Frequency with Strategically Timed Attack}%
        \label{fig:timed_attacks}
    \end{subfigure}
    \begin{subfigure}[b]{\textwidth}
        \hspace{0em}\begin{subfigure}[b]{\imgwidth\textwidth}
            \includegraphics[width=\textwidth]{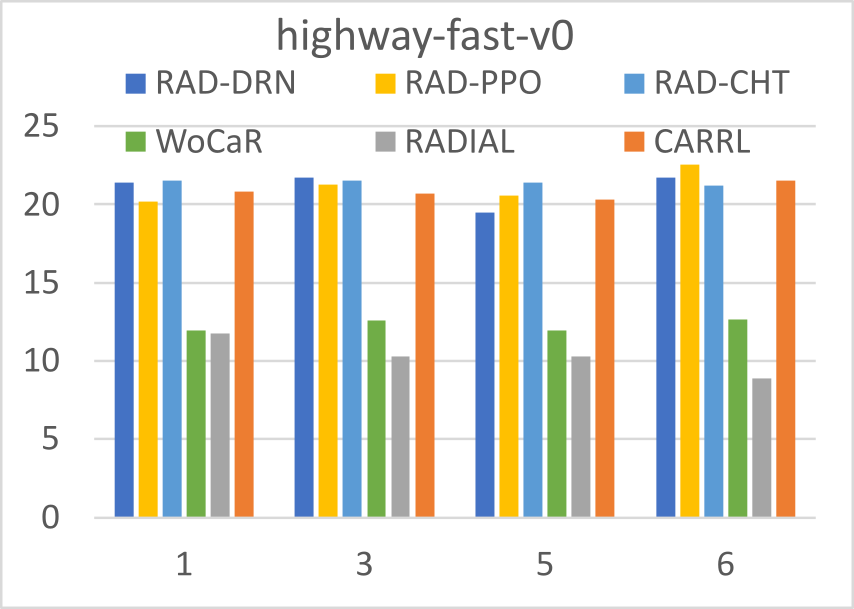}
            \label{fig:crit:hwy}
        \end{subfigure}%
        %\hfill
        \hspace{0em}\begin{subfigure}[b]{\imgwidth\textwidth}
            \includegraphics[width=\textwidth]{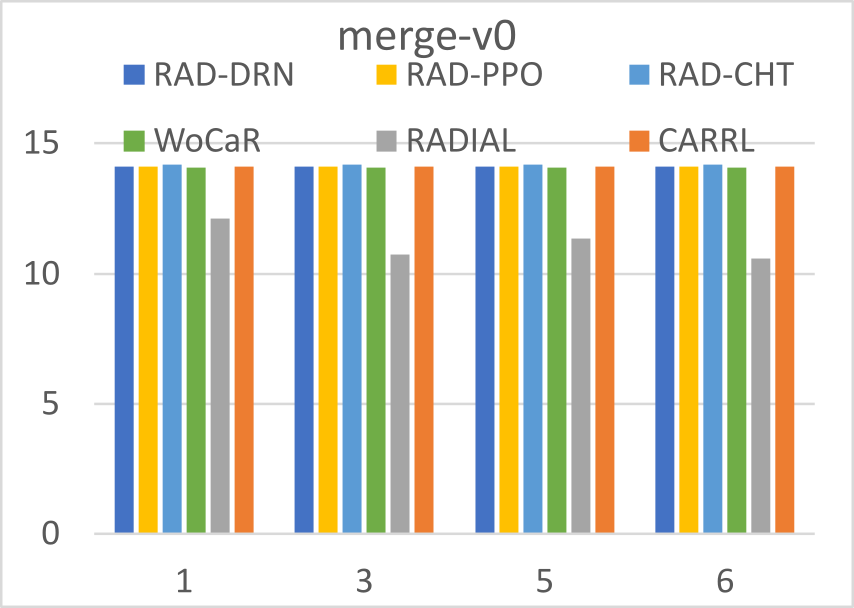}
            \label{fig:crit:mrg}
        \end{subfigure}%
        %\hfill
        \hspace{0em}\begin{subfigure}[b]{\imgwidth\textwidth}
            \includegraphics[width=\textwidth]{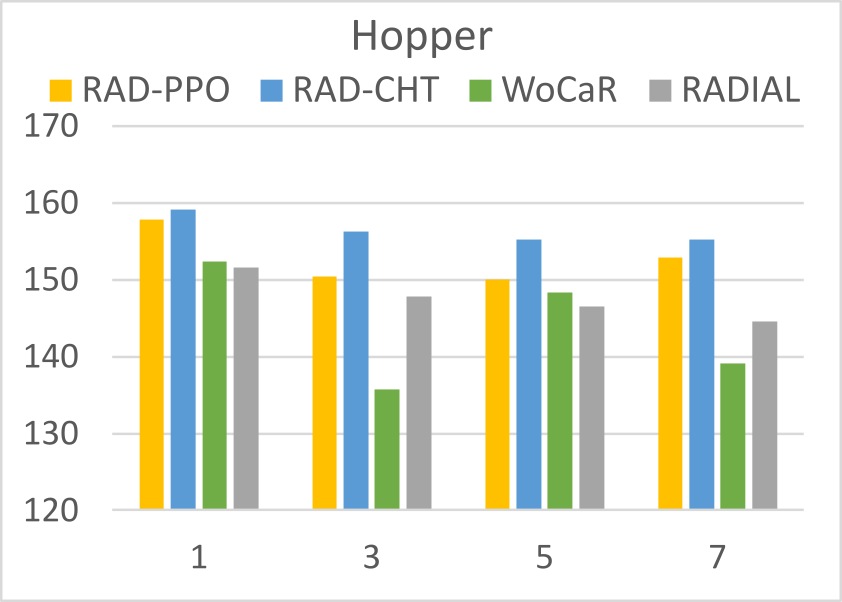}
            \label{fig:crit:hopper}
        \end{subfigure}%
        %\hfill
        \begin{subfigure}[b]{\imgwidth\textwidth}
            \includegraphics[width=\textwidth]{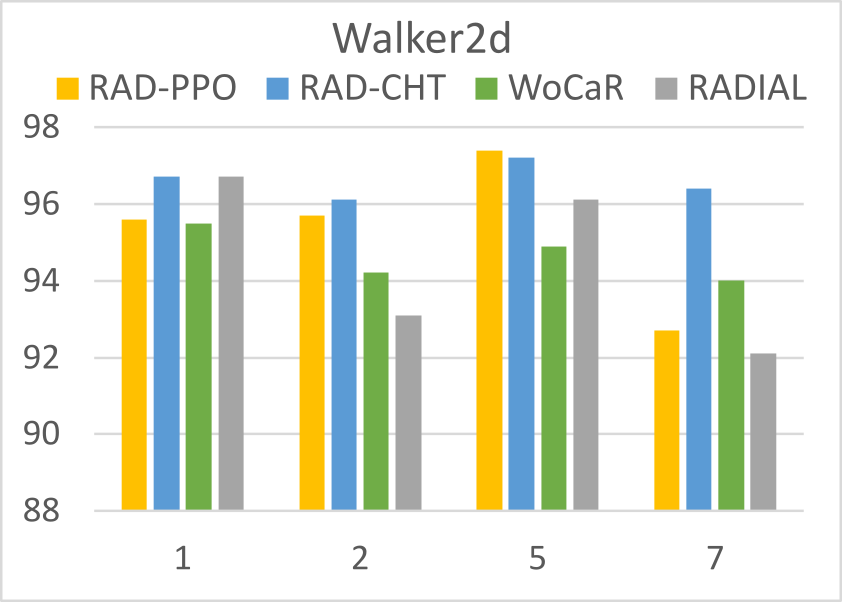}
            \label{fig:crit:walker}
        \end{subfigure}%
        %\hfill
        \caption{Score vs. Percentile Perturbation Magnitude with Critical Point Attack}%
        \label{fig:crit_attacks}
    \end{subfigure}
    \caption{The performance of robust RL methods against strategic adversaries. The y-axis represents the score and the x-axis represents the intensity of the attack.}
    \label{fig:multistep}
\end{figure*}
\section{Experiments}
%{The authors in the experiments take $|N(s_t)| = 10$ as the size of the neighborhood. How was this value chosen?
%Have experiments been done on the impact of increasing or decreasing this value?}
We provide empirical evidence to answer key questions:
\begin{itemize}
    \item How do RAD approaches (RAD-DRN, RAD-PPO, RAD-CHT) compare against leading methods for Adversarially Robust RL on well-known baselines from MuJoCo, Atari, and Highway libraries?
    \item For multi-step strategic attacks, how do RAD methods compare to leading defenses? 
    \item Does RAD mitigate the value/robustness trade-off present in maximin methods?
    \item How does the performance of our approaches degrade as the intensity of attacks (i.e., \# of attacks) is increased?
\end{itemize}

% \begin{figure}
%     \centering
%     \includegraphics{media/hwy.pdf}
%     \caption{The highway-fast-v0 environment, available at https://github.com/eleurent/highway-env \cite{highway-env}.}
%     \label{fig:highway_env}
% %    \Description{An image depicting the highway-fast-v0 environment, part of the highway-env package. A green ego-vehicle navigates between other blue vehicles on a four-lane highway. A green dotted line displays the ego-vehicle's current trajectory, aiming to remain in the rightmost lane.}
% \end{figure}

\subsection{Experimental Setup}
We evaluate our proposed methods on the commonly used Atari and MuJoCo domains \cite{Bellemare_2013:Atari, todorov2012mujoco}, and a suite of discrete-action self-driving tasks~\cite{highway-env}. For the driving environments, each observation feature of the agent corresponds to the kinematic properties (coordinates, velocities, and angular headings) of the ego-vehicle and nearby vehicles. The task for each environment is to maximize the time spent in the fast lane while avoiding collisions. The road configurations for the \textit{highway-fast-v0, merge-v0, roundabout-v0, and intersection-v0} tasks are, in order: a straight multi-lane highway, a two-lane highway with a merging on-ramp, a multi-lane roundabout, and a two-lane intersection. We use a standard training setup seen in \cite{oikarinen2021robust, liang2022efficient}, detailed in Appendix C.

We compare RAD-DRN, RAD-PPO, and RAD-CHT to the following baselines: vanilla DQN \cite{mnih2013} and PPO \cite{ppo}; a simple but robust minimax method, CARRL \cite{everett:carrl}; a leading regularization approach, RADIAL \cite{oikarinen2021robust}; and the current SoTA defense, WocaR \cite{liang2022efficient}. We test all methods against both trained policy adversaries and gradient attacks, as well as the Strategically-Timed attack and Critical-Point attack, two leading multi-step strategic attacks \cite{lin:tactics2017,Sun_Zhang_Xie_Ma_Zheng_Chen_Liu_2020}. For RADIAL and WocaR, we use the best-performing methods for each environment (their proposed DQNs for discrete actions, and PPOs for MuJoCo). 

%Full Results Table
\iftrue
\begin{table}[t!]
    \caption{Results on Highway. Each row shows the mean scores of each RL method against different attacks. Further tasks are shown in Appendix B.}
\centering
\label{tab:hwyresults}
\begin{tabular}{ |p{1.65cm}|p{1.85cm}|p{1.9cm}|p{1.85cm}| }
 \hline
 Algorithm & Unperturbed & WC Policy & PGD, $\epsilon\!\!=\!\frac{3}{255}$ \\
 \hline
 \multicolumn{4}{|c|}{highway-fast-v0} \\
 \hline
 DQN   & 24.91$\pm$20.27    &3.68$\pm$35.41     &15.71$\pm$13\\
 PPO &22.8$\pm$5.42         & 13.63$\pm$19.85   &15.21$\pm$16.1 \\
  CARRL & 24.4$\pm$1.10       & 4.86$\pm$15.4     & 12.43$\pm$3.4\\
 RADIAL& {28.55$\pm$0.01}  &2.42$\pm$1.3 &14.97$\pm$3.1 \\
 WocaR  & 21.49$\pm$0.01       & 6.15$\pm$0.3      & 6.19$\pm$0.4\\
 %DRN&   22.04$\pm$ 7.68     &18.04$\pm$9.01     &17.97$\pm$15 \\
 RAD-DRN    & 24.85$\pm$0.01   & \textbf{22.65$\pm$0.02} &{18.8$\pm$24.6} \\
 RAD-PPO    & 21.01$\pm$1.23   & 20.59$\pm$4.10     &{20.02$\pm$0.01} \\
 RAD-CHT    & 21.83 $\pm$ 0.35	& 21.1 $\pm$ 0.24	& \textbf{21.48 $\pm$ 1.8} \\

 \hline
\end{tabular}
\end{table}
\fi

\begin{table}[t!]
    \caption{Results on MuJoCo.}
\centering
\label{tab:mujocoresults}
\begin{tabular}{ |p{1.6cm}|p{1.8cm}|p{1.95cm}|p{1.75cm}| }
 \hline
 Algorithm& Unperturbed & MAD $\epsilon\!\!=\!0.15$ & PGD $\epsilon\!\!=\!\frac{5}{255}$ \\
 \hline
 \multicolumn{4}{|c|}{Hopper} \\
 \hline
 PPO &2741 $\pm$ 104        & 970$\pm$19   &36$\pm$156 \\
 RADIAL& {3737$\pm$75}  &2401$\pm$13 & 3070$\pm$31 \\
 WocaR  & 3136$\pm$463       &  1510 $\pm$ 519   & 2647 $\pm$310\\
 RAD-PPO   & 3473$\pm$23   & {2783$\pm$325}     &\textbf{3110$\pm$30} \\
 RAD-CHT    & 3506$\pm$377& 	\textbf{2910$\pm$ 699} & 3055 $\pm$152 \\

 \hline
 % \hline
 %Algorithm& Unperturbed & WC Policy & PGD $\epsilon\!\!=\!\frac{5}{255}$ \\
 \hline
 \multicolumn{4}{|c|}{HalfCheetah} \\
 \hline
 PPO &{5566 $\pm$ 12}        & 1483$\pm$20   &-27$\pm$1308 \\
 RADIAL& 4724$\pm$76  &4008$\pm$450 &{3911$\pm$129} \\
 WocaR  & 3993$\pm$152     &  3530$\pm$458  & 3475$\pm$610\\
 RAD-PPO    & 4426$\pm$54   & \textbf{4240$\pm$4}     & \textbf{4022$\pm$851}\\
 RAD-CHT    & 4230$\pm$140& 4180$\pm$37 & 3934$\pm$486 \\

 \hline
  % \hline
 %Algorithm& Unperturbed & WC Policy & PGD $\epsilon\!\!=\!\frac{5}{255}$ \\
 \hline
 \multicolumn{4}{|c|}{Walker2d} \\
 \hline
 PPO &3635 $\pm$ 12        & 680$\pm$1570   &730$\pm$262 \\
 RADIAL& {5251$\pm$10} &3895$\pm$128 &3480$\pm$3.1 \\
 WocaR  & 4594$\pm$974       & {3928$\pm$1305}      & 3944$\pm$508\\
 RAD-PPO    & 4743$\pm$78& 3922$\pm$426    & \textbf{4136$\pm$639}\\
 RAD-CHT    &  4790$\pm$61  & \textbf{4228$\pm$539} &  4009$\pm$516\\

 \hline

\end{tabular}
\end{table}

\begin{table}[t!]
    \caption{Results on Atari, with the same metrics as Table \ref{tab:hwyresults}. Additional results in Appendix B.}
\centering
\label{tab:atariresults}
\begin{tabular}{ |p{1.7cm}|p{1.8cm}|p{1.8cm}|p{1.8cm}| }
 \hline
 Algorithm& Unperturbed & WC Policy & PGD $\epsilon\!\!=\!\frac{5}{255}$ \\
 \hline
 \multicolumn{4}{|c|}{Pong} \\
 \hline
 %DQN &  {21.0$\pm$0} &  -21.0$\pm$0.0                & -21.0$\pm$0.0\\
 PPO &{21.0$\pm$0}   & -20.0$\pm$ 0.07   &-19.0$\pm$1.0 \\
 CARRL &  13.0 $\pm$1.2&    11.0$\pm$0.010             & 6.0$\pm$1.2 \\
 RADIAL& {21.0$\pm$0}&11.0$\pm$2.9 &\textbf{21.0$\pm$ 0.01} \\
 WocaR  & {21.0$\pm$0} &   18.7 $\pm$0.10  & 20.0 $\pm$ 0.21\\
 RAD-DRN & {21.0$\pm$0}&  14.0 $\pm$ 0.04   & 14.0 $\pm$ 2.40 \\
 RAD-PPO    & {21.0$\pm$0}& \textbf{20.1$\pm$1.0}     &20.8$\pm$0.02 \\
 %RAD-CHT    & & 	&  \\

 \hline
  % \hline
% Algorithm& Unperturbed & WC Policy & PGD $\epsilon\!\!=\!\frac{5}{255}$ \\
\iffalse
 \hline
 \multicolumn{4}{|c|}{Freeway} \\
 \hline
 %DQN &  {33.9$\pm$0.10}& 0$\pm$0.0  & 2$\pm$1.10 \\
 PPO &29 $\pm$ 3.0       & 4 $\pm$ 2.31   &2$\pm$2.0 \\
 CARRL &    18.5$\pm$0.0   &   19.1 $\pm$1.20   &  15.4$\pm$0.22\\
 RADIAL& 33.2$\pm$0.19  &29.0$\pm$1.1 &24.0$\pm$0.10 \\
 WocaR  & 31.2$\pm$0.41       & 19.8$\pm$3.81   & 28.1$\pm$3.24\\
 RAD-DRN &  33.2$\pm$0.18&  \textbf{30.0$\pm$0.23} & 27.7$\pm$1.51\\
 RAD-PPO    & 33.0 $\pm$0.12 & \textbf{30.0$\pm$0.10}     & \textbf{29 $\pm$0.12}\\
 %RAD-CHT    & & &  \\

 \hline
 \fi
  % \hline
% Algorithm& Unperturbed & WC Policy & PGD $\epsilon\!\!=\!\frac{5}{255}$ \\
 \hline
 \multicolumn{4}{|c|}{BankHeist} \\
 \hline
 %DQN & 1325$\pm$5 &    0$\pm$0  & 0.4$\pm$55 \\
 PPO &1350$\pm$0.1         & 680$\pm$419   &0$\pm$116 \\
 CARRL &  849$\pm$0& 830$\pm$32   & 790$\pm$110\\
 RADIAL& {1349$\pm$0 }&997$\pm$3 &1130$\pm$6 \\
 WocaR  & 1220$\pm$0       & 1207$\pm$39      & 1154$\pm$94\\
 RAD-DRN & 1340$\pm$0 &  1170$\pm$42  & 1211$\pm$56 \\
 RAD-PPO    & 1340$\pm0$& \textbf{1301$\pm$8}     & \textbf{1335$\pm$52}\\
 %RAD-CHT    & & &  \\

 \hline
  % \hline
  \iffalse
 %Algorithm& Unperturbed & WC Policy & PGD $\epsilon\!\!=\!\frac{5}{255}$ \\
 \hline
 \multicolumn{4}{|c|}{RoadRunner} \\
 \hline
 %DQN & 43380 $\pm$860&   1193 $\pm$259               &  3940 $\pm$ 92 \\
 PPO &42970$\pm$210         & 18309$\pm$485   &10003$\pm$521 \\
 CARRL & 26510$\pm$20 & 24480$\pm$ 200 & 22100$\pm$370\\
 RADIAL& {44501$\pm$1360} &23119$\pm$1100 &24300$\pm$1315 \\
 WocaR  & 44156$\pm$2270       & 25570$\pm$390      & 12750$\pm$405\\
 RAD-DRN & 42900$\pm$1020 &  29090$\pm$440  & 27150$\pm$505\\
 RAD-PPO    & 42805$\pm$240& \textbf{33960$\pm$360}     & \textbf{33990$\pm$403}\\
 %RAD-CHT    & & &  \\

 \hline
\fi
\end{tabular}
\end{table}

\subsection{Worst Case Policy Attack} \label{AdvTrain}
Each Worst-Case Policy (WC) adversary is a Q-learning agent that minimizes the overall return of the targeted agents by learning perturbed observations from a neighborhood distribution of surrounding states $N(s_t)$. Note that $N(s_t)$ is not necessarily equal to the training neighborhood that DRN samples; in our experiments, we increase the attack neighborhood radius by twenty percent (approximately). During the training of the adversary, it is permitted to perturb the victim's observations at every step. Each tested method has a corresponding WC Policy attacker. In the Highway environments, the perturbations correspond to a small shift in the position of a nearby vehicle; in the Atari domains, the input tensor is shifted up or down several pixels. For MuJoCo, we instead apply the Maximal Action Difference attack with $\epsilon=0.15$ \cite{zhangsadqn}.

\subsection{Results}
In each table, we report the mean return over 50 random seeds. {The most robust score is shown in \textbf{boldface}. }
%In our results, reported values are taken as an average across 100 runs: we run each environment with 50 different random seeds, completing each seeded environment twice (more if $t_{adv}$ is higher).

\noindent\textbf{Highway Domains: }We report scores under PGD and WC attacks in Table \ref{tab:hwyresults}. We observe that although the unperturbed performance of RAD- methods is lower than that of the vanilla solutions, all three approaches (RAD-DRN, RAD-PPO, and RAD-CHT) are extremely robust under different attacks and have higher performance than all other robust approaches in all of the Highway environments.  We see that even though maximin methods (CARRL, WocaR) are clearly suited for some scenarios, RAD- methods still outperform them.

%For the tasks in this section, we trained RAD- approaches both from scratch (training value networks and regret networks simultaneously) and with a pre-trained value network. Empirically, we find that the reported scores are attainable via both setups, with training times for both setups being twice that of a vanilla DQN (the actual number varies per task). 

\noindent\textbf{MuJoCo Domains: }Table \ref{tab:mujocoresults} reports the results on MuJoCo, playing three commonly tested environments seen in comparison literature \cite{oikarinen2021robust,liang2022efficient}. As the MuJoCo domains have continuous actions, we exclude value iteration-based methods (DQN, CARRL, and RAD-DRN). We directly integrated our proposed techniques into the implementation from \cite{liang2022efficient}. Our assessment covers the MAD attack \cite{zhangsadqn} with a testing $\epsilon=0.15$, which marks the value at which we start observing deviations in the returns of the robustly trained agents. In addition, we subject the agents to PGD attacks with $\epsilon = \frac{5}{255}$.

Even in these problem settings, while RAD approaches do not provide the highest unperturbed performance (but are reasonably close), they (RAD-PPO and RAD-CHT) provide the best overall performance of all approaches. 

\noindent\textbf{Atari Domains: }We report results on four Atari domains used in preceding works \cite{liang2022efficient,oikarinen2021robust} in Table \ref{tab:atariresults}. RAD-PPO outperforms other robust methods under nearly all attacks. RAD-DRN performs similarly to RAD-PPO, though faces some limitations in scenarios where rewards are sparse or mostly similar, such as in \textit{Pong}. Regret measures the difference between outcomes, so the quality of training samples degrades when rewards are the same regardless of action. Interestingly, RAD-PPO does not suffer in this way, which we attribute to the superiority of the PPO framework over value iterative methods.\par
We exclude RAD-CHT from the Atari experiments, as the memory requirements to simultaneously hold models for all levels are outside the limitations of our hardware. 

\noindent\textbf{Strategic Attacks: }Unlike previous works in robust RL, we also test our methods against attackers with a longer planning horizon than the above-mentioned greedy adversaries. In Figure \ref{fig:multistep}, we test under a) the Strategically Timed Attack \cite{lin:tactics2017} and b) the Critical Point attack \cite{Sun_Zhang_Xie_Ma_Zheng_Chen_Liu_2020}. We observe that across all domains, regret defense outperforms other robustness methods. Particularly, regularization methods fail with increasing rates as the strength or frequency of the attacks increases, even as maximin methods (CARRL) retain some level of robustness. This is one of the main advantages of our proposed methods, as the resulting policies seek robust trajectories to occupy rather than robust single-step action distributions.

%%%%%%%%%%%%%%%%%%%%%%%%%%%%%%%%%%%%%%%%%%%%%%%%%%%%%%%%%%%%%%%%%%%%%%%%

\section{Conclusion}
% {The conclusion section is a bit short. Limitations of the proposed approach and some more detailed future
% directions should be included.}
We show that regret can be used to increase the robustness of RL to adversarial observations, even against stronger or previously unseen attackers. We propose an approximation of regret, CCER, and demonstrate its usefulness in proactive value iterative and policy gradient methods, and reactive training under CHT. Our results on a wide variety of problems (more in Appendix B) show that regret-based defense significantly improves robustness against strong observation attacks from both greedy and strategic adversaries. Specifically, RAD-PPO performs the best on average across all (including Appendix B) of our experimental results. 

%%%%%%%%%%%%%%%%%%%%%%%%%%%%%%%%%%%%%%%%%%%%%%%%%%%%%%%%%%%%%%%%%%%%%%%%
\begin{acks}
    This research/project is supported by the National Research Foundation Singapore and DSO National Laboratories under the AI Singapore Programme (AISG Award No: AISG2-RP-2020-017)
\end{acks}

%%% The next two lines define, first, the bibliography style to be 
%%% applied, and, second, the bibliography file to be used.
\newpage
\balance
\bibliographystyle{ACM-Reference-Format} 
\bibliography{main}

%%% Appendix

\newpage

\onecolumn
\section*{Appendix A: Proofs}
\subsection*{Proof of Proposition 1}
\begin{proof}
Since $\pi^\bot$ is optimal for all states from time step $t+1$:
\begin{align}
     \delta_{\texttt{CCER}}^{\pi^\bot_{[t+1,H]}}(z_{t+1}) &\leq \delta_{\texttt{CCER}}^{\pi_{[t+1,H]}}(z_{t+1}) \label{eq:opt}
\end{align}
Therefore, we obtain:
\begin{align}
    &\delta_\texttt{CCER}^{\left<\pi_t, \pi^\bot_{[t+1,H]}\right>}\!(z_t) \!=\! R(z_t,\pi_t(z_t)) - \min\nolimits_{s_t \in N({{z}_t})} \!R(s_t, \pi_t(z_t))  \nonumber\\
    &\qquad\qquad\qquad\qquad +\gamma \mathbb{E}_{z_{t+1}}\Big[\delta_{\texttt{CCER}}^{\pi^\bot_{[t+1,H]}}(z_{t+1})\Big] \nonumber\\
    &\qquad\quad\leq  R(z_t,\pi_t(z_t)) - \min\nolimits_{s_t \in N({{z}_t})} R(s_t, \pi_t(z_t)) \label{ineq.1} \\
    &\qquad\qquad\qquad\qquad + \gamma E_{z_{t+1}}[\delta_{\texttt{CCER}}^{\pi_{[t+1,H]}}(z_{t+1})] \nonumber\\
     %&\leq  R(z_t,\pi_2(z_t)) - \min_{s_t \in N({{z}_t})} R(s_t, \pi_2(z_t))  \label{ineq.2}\\
    %&\qquad\qquad +\gamma E_{z_{t+1}}[\delta_{\texttt{CCER}}^{\pi_2}(z_{t+1})] \nonumber\\
    &\qquad\quad\leq \delta_{\texttt{CCER}}^{\pi_t,\pi_{[t+1,H]}}(z_{t+1})
\end{align}
% \begin{align}
%     Q^{CCER}_{\pi_1}(z_{t+1}) &\leq Q^{CCER}_{\pi_2}(z_{t+1}) \label{eq:opt}
%     \intertext{CCER for $\pi_1$ is given by}
%     Q^{CCER}_{\pi_1}(z_t) &= R(z_t,\pi_1(z_t)) - \min_{s_t \in N({{s}_t})} R(s_t, \pi_1(z_t)) + \nonumber\\
%     &\hspace{0.2in} \gamma E_{z_{t+1}}[Q^{CCER}_{\pi_1}(z_{t+1})] \nonumber\\
%     \intertext{Utilizing Equation~\ref{eq:opt}}
%     &\leq  R(z_t,\pi_1(z_t)) - \min_{s_t \in N({{s}_t})} R(s_t, \pi_1(z_t)) + \nonumber\\
%     &\hspace{0.2in} \gamma E_{z_{t+1}}[Q^{CCER}_{\pi_2}(z_{t+1})] \nonumber\\
%     \intertext{From proposition statement, $\pi_1(z_t) = \pi_2(z_t) \; \forall z_t$, therefore}
%     &\leq  R(z_t,\pi_2(z_t)) - \min_{s_t \in N({{s}_t})} R(s_t, \pi_2(z_t)) + \nonumber\\
%     &\hspace{0.2in} \gamma E_{z_{t+1}}[Q^{CCER}_{\pi_2}(z_{t+1})] \nonumber\\
%     &\leq Q^{CCER}_{\pi_2}(z_t) 
% \end{align}
 which concludes our proof.
\end{proof}
\subsection*{Proof of Proposition 2}
\begin{proof} We have: 
\begin{align*}
    &\frac{\partial \delta_{\texttt{CCER}}^{\pi_\theta}(z)}{\partial\theta} = \frac{\partial}{\partial\theta}\sum\nolimits_a \pi_\theta(z,a)\delta_{\texttt{CCER}}^{\pi_\theta} (z,a) \\
    &=\!\!\sum\nolimits_a \!\!\Big[\frac{\partial\pi_\theta(z,a)}{\partial\theta}\delta_{\texttt{CCER}}^{\pi_\theta} (z,a) \!+\! \pi_\theta(z,a)\!\Big[\frac{\partial}{\partial\theta}\delta_{\texttt{CCER}}^{\pi_\theta}(z,a)\Big]\Big]\\
    &=\!\!\sum\nolimits_a \!\!\Big[\frac{\partial\pi_\theta(z,a)}{\partial\theta}\delta_{\texttt{CCER}}^{\pi_\theta} (z,a) \!+\! \pi_\theta(z,a)\!\Big[ \!\sum\nolimits_{z'}\!\! \gamma\frac{\partial}{\partial\theta} \delta_{\texttt{CCER}}^{\pi_\theta}(z')\Big]\Big]
\end{align*}
After unrolling this equation, and adjusting terms, we obtain the following overall gradient computation: 
\begin{align}%\label{eq:policygradient}
    \frac{\partial \delta_{\texttt{CCER}}^{\pi_\theta}(z)}{\partial\theta} &\!=\! \sum\nolimits_z \!\!P(z|\pi_\theta) \!\sum\nolimits_a\!\! \frac{\partial\pi_\theta(z,a)}{\partial\theta}\delta_{\texttt{CCER}}^{\pi_\theta} (z,a)
\end{align}
which concludes our proof.
\end{proof}
\subsection*{Proof of Proposition 3}
\begin{proof}
     We elaborate the objective function of the adversary as an expectation over all possible trajectories as follows:
     \begin{align*}
         & V_{\mu_{\theta}} = \sum\nolimits_iP^{(k)}(i)\sum\nolimits_{\tau} P^{(i)}(\tau) R(\tau)
     \end{align*}
     where the probability of a trajectory is computed as:
     \begin{align*}
         & P^{(i)}(\tau) = \prod\nolimits_t \mu_{\theta}(z_t\mid s_t) \pi^{(i)}(a_t\mid z_t)P(s_{t+1}\mid s_t, a_t)
     \end{align*}
     Therefore, we can compute the gradient $\nabla_{\theta}V_{\mu_{\theta}}$ as follows:
     \begin{align*}
          &\sum\nolimits_iP^{(k)}(i)\sum\nolimits_{\tau}R(\tau)\nabla_{\tau} P^{(i)}(\tau) \\
         &=\sum\nolimits_iP^{(k)}(i)\sum\nolimits_{\tau}P^{(i)}(\tau)R(\tau)\nabla_{\theta} \log P^{(i)}(\tau)\\
         & = \sum\nolimits_iP^{(k)}(i)\sum\nolimits_{\tau}P^{(i)}(\tau)R(\tau)\sum\nolimits_t\nabla_{\theta} \log \mu_{\theta}(z_t| s_t)\\
         &=\mathbb{E}_{i\sim P^{(k)}(i), \tau\sim (\mu_{\theta}, \pi^{(i)}, T)} \Big[R(\tau) \sum\nolimits_t \nabla_\theta\log \mu_{\theta}(z_t|s_t)\Big]
     \end{align*}
     which concludes our proof.
 \end{proof}

\section*{Appendix B: More Results}
\subsection{Main Results Extended}
Here we include our experimental results in their entirety; Some were removed for brevity. 

\begin{table}[H]
    \caption{Results on MuJoCo. Each row shows the mean scores of each RL method against different attacks.}
\centering
\label{tab:mujocoresults}
\begin{tabular}{ |p{2cm}|p{2cm}|p{2cm}|p{2cm}| }
 \hline
 Algorithm& Unperturbed & MAD $\epsilon\!\!=\!0.15$ & PGD $\epsilon\!\!=\!\frac{5}{255}$ \\
 \hline
 \multicolumn{4}{|c|}{Hopper} \\
 \hline
 PPO &2741 $\pm$ 104        & 970$\pm$19   &36$\pm$156 \\
 RADIAL& \textbf{3737$\pm$75}  &2401$\pm$13 & 3070$\pm$31 \\
 WoCaR  & 3136$\pm$463       &  1510 $\pm$ 519   & 2647 $\pm$310\\
 RAD-PPO   & 3473$\pm$23   & {2783$\pm$325}     &\textbf{3110$\pm$30} \\
 RAD-CHT    & 3506$\pm$377& 	\textbf{2910$\pm$ 699} & 3055 $\pm$152 \\

 \hline
 % \hline
 %Algorithm& Unperturbed & WC Policy & PGD $\epsilon\!\!=\!\frac{5}{255}$ \\
 \hline
 \multicolumn{4}{|c|}{HalfCheetah} \\
 \hline
 PPO &\textbf{5566 $\pm$ 12}        & 1483$\pm$20   &-27$\pm$1308 \\
 RADIAL& 4724$\pm$76  &4008$\pm$450 &{3911$\pm$129} \\
 WoCaR  & 3993$\pm$152     &  3530$\pm$458  & 3475$\pm$610\\
 RAD-PPO    & 4426$\pm$54   & \textbf{4240$\pm$4}     & \textbf{4022$\pm$851}\\
 RAD-CHT    & 4230$\pm$140& 4180$\pm$37 & 3934$\pm$486 \\

 \hline
  % \hline
 %Algorithm& Unperturbed & WC Policy & PGD $\epsilon\!\!=\!\frac{5}{255}$ \\
 \hline
 \multicolumn{4}{|c|}{Walker2d} \\
 \hline
 PPO &3635 $\pm$ 12        & 680$\pm$1570   &730$\pm$262 \\
 RADIAL& \textbf{5251$\pm$10} &3895$\pm$128 &3480$\pm$3.1 \\
 WoCaR  & 4594$\pm$974       & {3928$\pm$1305}      & 3944$\pm$508\\
 RAD-PPO    & 4743$\pm$78& 3922$\pm$426    & \textbf{4136$\pm$639}\\
 RAD-CHT    &  4790$\pm$61  & \textbf{4228$\pm$539} &  4009$\pm$516\\

 \hline

\end{tabular}
\end{table}

\begin{table}[H]
    \caption{Results on Highway. Each row shows the mean scores of each RL method against different attacks. }
\centering
\label{tab:hwyresults}
\begin{tabular}{ |p{2cm}|p{2cm}|p{2cm}|p{2.2cm}| }
 \hline
 Algorithm & Unperturbed & WC Policy & PGD, $\epsilon\!\!=\!\frac{3}{255}$ \\
 \hline
 \multicolumn{4}{|c|}{highway-fast-v0} \\
 \hline
 DQN   & 24.91$\pm$20.27    &3.68$\pm$35.41     &15.71$\pm$13\\
 PPO &22.8$\pm$5.42         & 13.63$\pm$19.85   &15.21$\pm$16.1 \\
  CARRL & 24.4$\pm$1.10       & 4.86$\pm$15.4     & 12.43$\pm$3.4\\
 RADIAL& \textbf{28.55$\pm$0.01}  &2.42$\pm$1.3 &14.97$\pm$3.1 \\
 WoCaR  & 21.49$\pm$0.01       & 6.15$\pm$0.3      & 6.19$\pm$0.4\\
 %DRN&   22.04$\pm$ 7.68     &18.04$\pm$9.01     &17.97$\pm$15 \\
 RAD-DRN    & 24.85$\pm$0.01   & \textbf{22.65$\pm$0.02} &{18.8$\pm$24.6} \\
 RAD-PPO    & 21.01$\pm$1.23   & 20.59$\pm$4.10     &{20.02$\pm$0.01} \\
 RAD-CHT    & 21.83 $\pm$ 0.35	& 21.1 $\pm$ 0.24	& \textbf{21.48 $\pm$ 1.8} \\

 \hline
 \hline
 \multicolumn{4}{|c|}{merge-v0} \\
 \hline
 DQN   & 14.83$\pm$0.0      &6.97$\pm$0.03  & 10.82$\pm$2.85 \\
 PPO   &11.08$\pm$0.01       &10.2$\pm$0.02   &10.42$\pm$0.95\\
 CARRL & 12.6$\pm$0.01     & 12.6$\pm$0.01     & 12.02$\pm$0.01\\
 RADIAL    & {14.86$\pm$0.01} & 11.29$\pm$0.01   & 11.04$\pm$0.91 \\
 WoCaR  & \textbf{14.91$\pm$0.04} & 12.01$\pm$0.28 & 11.71$\pm$0.21\\
% DRN   &   11.02$\pm$0.0    &   10.19$\pm$0.0  & 10.41$\pm$2.01\\
  RAD-DRN    & {14.86$\pm$0.01}  &{13.23$\pm$0.01}   &{11.77$\pm$1.07}\\
  RAD-PPO & 13.91$\pm$0.01        &  {13.90$\pm$0.01} &11.72$\pm$0.01\\
  RAD-CHT   & 14.7 $\pm$ 0.01	&\textbf{14.7 $\pm$ 0.12}&\textbf{13.01 $\pm$ 0.32} \\

 \hline

 \hline
 \multicolumn{4}{|c|}{roundabout-v0} \\
 \hline
 DQN   & 9.55$\pm$0.081    &8.97$\pm$0.26           & 9.01$\pm$1.9\\
 PPO   & 10.33$\pm$0.40      & 9.23$\pm$0.69        & 9.09$\pm$1.35\\
  CARRL & 9.75$\pm$0.01      & \textbf{9.75$\pm$0.01} & 5.92$\pm$0.12\\
  RADIAL    & 10.29$\pm$0.01  & 5.33$\pm$7e-31      & 8.77$\pm$2.4    \\%&19.40$\pm$0.0\\
 WoCaR  & 6.75$\pm$2.5      & 6.05$\pm$0.14         & 6.48$\pm$2.7\\
 %DRN   &   9.06$\pm$0.010   & 9.06$\pm$0.002        & 8.95 $\pm$1.9 \\
 DRN   &   \textbf{10.39$\pm$0.01}  &9.33$\pm$3e-30 & \textbf{9.18$\pm$2.1} \\
   RAD-PPO & 9.22$\pm$0.3 &  8.98$\pm$0.3 &9.11$\pm$0.3\\
   RAD-CHT   & 9.79 $\pm$ 0.02&	9.12 $\pm$ 0.01        &9.5 $\pm$ 0.9\\

 \hline
 \hline
 \multicolumn{4}{|c|}{intersection-v0} \\
 \hline
 DQN   & 9.9$\pm$0.49    &1.31$\pm$2.85 & 8.61$\pm$0.1 \\
 PPO   & 9.26$\pm$7.6    & 3.62$\pm$11.63 & 6.75$\pm$12.93\\
RADIAL   &   \textbf{10.0$\pm$0}  &   2.4$\pm$5.1  & 9.61$\pm$0.1 \\
 WoCaR  & \textbf{10.0$\pm$0.05} & 9.47$\pm$0.3 & 3.26$\pm$0.4\\
 CARRL & 8.0$\pm$0 & 7.5$\pm$0 & 9.0$\pm$0.1\\
 %DRN   & \textbf{10.0$\pm$0}    & {9.64$\pm$0}   & 9.64$\pm$0.1 \\
 DRN   &   \textbf{10.0$\pm$0}  & \textbf{10.0$\pm$0 } & \textbf{9.68$\pm$0.1} \\
   RAD-PPO & 9.85$\pm$1.2 & 9.71$\pm$2.3 &9.62$\pm$0.1\\
   RAD-CHT  & 4.52 $\pm$ 1.9	& 4.50 $\pm$ 2.5	& 4.67 $\pm$ 3.1 \\ 
\hline

 %\hline
 %\hline
 %\multicolumn{4}{|c|}{CartPole-v0} \\
 %\hline
 %DQN   & \textbf{200$\pm$0}    &107$\pm$3e3   & 49.17$\pm$3.9e3 \\%&56.9\\
 %PPO   &{196.7$\pm$2.1}     &43.22$\pm$3.1e3 & 65.94$\pm$3.1e3  \\%&75.47\\
 %RADIAL    & \textbf{200$\pm$0} & 123.15$\pm$2.2e2 & 51.53$\pm$1.5e3 \\%&19.40$\pm$0.0\\
 % CARRL & \textbf{200$\pm$0} &  160$\pm$33 &26.02$\pm$23\\
 %DRN   & \textbf{200$\pm$0}    & 133$\pm$28  & 68.6$\pm$5.3e3 \\%&200\\
 % DRN+    & \textbf{200$\pm$0}     & \textbf{191.2 $\pm$102} &\textbf{151.43$\pm$1e3}\\
 % RAD-PPO & 194.1$\pm$ 8.0 &  183.9$\pm$89.1 & 128$\pm$91.5\\
%\hline
% \hline
% \multicolumn{4}{|c|}{LunarLander-v2} \\
% \hline
% DQN   & \textbf{200$\pm$0}    & -3.19$\pm$19.5 & -14.07$\pm$5.1 \\%&56.9\\
% PPO  & \textbf{200$\pm$0}    &  -0.89$\pm$10.1  & 45.2$\pm$4.3  \\%&75.47\\
% RADIAL    & \textbf{200$\pm$0}  & 34.7$\pm$3.7 & 17.91$\pm$1.2 \\%&19.40$\pm$0.0\\
% DRN+   & \textbf{200$\pm$0}  & \textbf{92.95$\pm$1.3} &\textbf{56.57$\pm$3.5} \\%&200\\
 % RAD-PPO    & -  & -     & -  \\
%\hline
\end{tabular}
\end{table}

\begin{table}[H]
    \caption{Results on Atari. Each row shows the mean scores of each RL method against different attacks. }
\centering
\label{tab:atariresults}
\begin{tabular}{ |p{2cm}|p{2cm}|p{2cm}|p{2.2cm}| }
 \hline
 Algorithm& Unperturbed & WC Policy & PGD $\epsilon\!\!=\!\frac{5}{255}$ \\
 \hline
 \multicolumn{4}{|c|}{Pong} \\
 \hline
 DQN &  \textbf{21.0$\pm$0} &  -21.0$\pm$0.0                & -21.0$\pm$0.0\\
 PPO &\textbf{21.0$\pm$0}   & -20.0$\pm$ 0.07   &-19.0$\pm$1.0 \\
 CARRL &  13.0 $\pm$1.2&    11.0$\pm$0.010             & 6.0$\pm$1.2 \\
 RADIAL& \textbf{21.0$\pm$0}&11.0$\pm$2.9 &\textbf{21.0$\pm$ 0.01} \\
 WoCaR  & \textbf{21.0$\pm$0} &   18.7 $\pm$0.10  & 20.0 $\pm$ 0.21\\
 RAD-DRN & \textbf{21.0$\pm$0}&  14.0 $\pm$ 0.04   & 14.0 $\pm$ 2.40 \\
 RAD-PPO    & \textbf{21.0$\pm$0}& \textbf{20.1$\pm$1.0}     &20.8$\pm$0.02 \\
 %RAD-CHT    & & 	&  \\

 \hline
  % \hline
% Algorithm& Unperturbed & WC Policy & PGD $\epsilon\!\!=\!\frac{5}{255}$ \\
 \hline
 \multicolumn{4}{|c|}{Freeway} \\
 \hline
 DQN &  \textbf{33.9$\pm$0.10}& 0$\pm$0.0  & 2$\pm$1.10 \\
 PPO &29 $\pm$ 3.0       & 4 $\pm$ 2.31   &2$\pm$2.0 \\
 CARRL &    18.5$\pm$0.0   &   19.1 $\pm$1.20   &  15.4$\pm$0.22\\
 RADIAL& 33.2$\pm$0.19  &29.0$\pm$1.1 &24.0$\pm$0.10 \\
 WoCaR  & 31.2$\pm$0.41       & 19.8$\pm$3.81   & 28.1$\pm$3.24\\
 RAD-DRN &  33.2$\pm$0.18&  \textbf{30.0$\pm$0.23} & 27.7$\pm$1.51\\
 RAD-PPO    & 33.0 $\pm$0.12 & \textbf{30.0$\pm$0.10}     & \textbf{29 $\pm$0.12}\\
 %RAD-CHT    & & &  \\

 \hline
  % \hline
% Algorithm& Unperturbed & WC Policy & PGD $\epsilon\!\!=\!\frac{5}{255}$ \\
 \hline
 \multicolumn{4}{|c|}{BankHeist} \\
 \hline
 DQN & 1325$\pm$5 &    0$\pm$0  & 0.4$\pm$55 \\
 PPO &1350$\pm$0.1         & 680$\pm$419   &0$\pm$116 \\
 CARRL &  849$\pm$0& 830$\pm$32   & 790$\pm$110\\
 RADIAL& \textbf{1349$\pm$0 }&997$\pm$3 &1130$\pm$6 \\
 WoCaR  & 1220$\pm$0       & 1207$\pm$39      & 1154$\pm$94\\
 RAD-DRN & 1340$\pm$0 &  1170$\pm$42  & 1211$\pm$56 \\
 RAD-PPO    & 1340$\pm0$& \textbf{1301$\pm$8}     & \textbf{1335$\pm$52}\\
 %RAD-CHT    & & &  \\

 \hline
  % \hline
 %Algorithm& Unperturbed & WC Policy & PGD $\epsilon\!\!=\!\frac{5}{255}$ \\
 \hline
 \multicolumn{4}{|c|}{RoadRunner} \\
 \hline
 DQN & 43380 $\pm$860&   1193 $\pm$259               &  3940 $\pm$ 92 \\
 PPO &42970$\pm$210         & 18309$\pm$485   &10003$\pm$521 \\
 CARRL & 26510$\pm$20 & 24480$\pm$ 200 & 22100$\pm$370\\
 RADIAL& \textbf{44501$\pm$1360} &23119$\pm$1100 &24300$\pm$1315 \\
 WoCaR  & 44156$\pm$2270       & 25570$\pm$390      & 12750$\pm$405\\
 RAD-DRN & 42900$\pm$1020 &  29090$\pm$440  & 27150$\pm$505\\
 RAD-PPO    & 42805$\pm$240& \textbf{33960$\pm$360}     & \textbf{33990$\pm$403}\\
 %RAD-CHT    & & &  \\

 \hline

\end{tabular}
\end{table}

\subsection{Strategic Attacks} In Figure \ref{fig:multistep} we report the effect of multi-step strategic attacks against robustly trained RL agents. In addition to the results presented in the main paper, we also include results on Atari domains.  %Due to the way in which the multi-step attack strategies are computed (in particular, the $M \times N$ adversarial simulation steps of the Critical Point Attack, where $M$ dictates the frequency of the attack strategy and $N$ is the number of attack strategies to test), the long time horizons and complex state space of Atari image tasks result in exceedingly long simulation times under default parameters. We reduce the number of adversarial simulation steps as well as the attack frequency, making the attacks less effective overall. For the Highway and MuJoCo environments we use default simulation parameters suggested by the respective authors of the listed attacks.
\def\imgwidth{0.25}
\begin{figure*}
    \begin{subfigure}[b]{\textwidth}
        \hspace{0em}\begin{subfigure}[b]{\imgwidth\textwidth}
        %\centering
            \includegraphics[width=\textwidth]{media/_hwy_timed.png}%
            \label{fig:timed:hwy}
        \end{subfigure}%
        \hspace{0em}\begin{subfigure}[b]{\imgwidth\textwidth}%
        %\centering
            \includegraphics[width=\textwidth]{media/_mrg_timed.png}%
            \label{fig:timed:mrg}
        \end{subfigure}%
        \hspace{0em}\begin{subfigure}[b]{\imgwidth\textwidth}%
        %\centering
            \includegraphics[width=\textwidth]{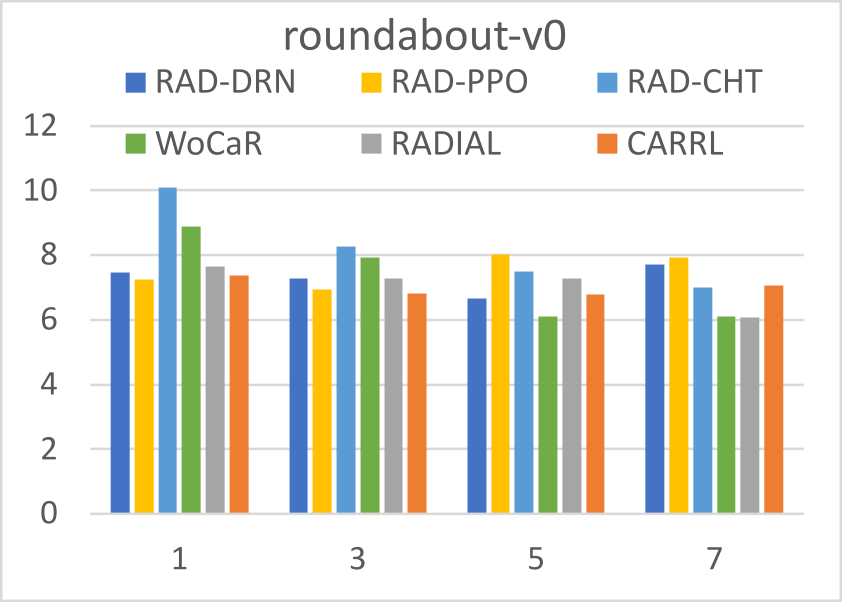}%
            \label{fig:timed:rdb}
        \end{subfigure}%
        \hspace{0em}\begin{subfigure}[b]{\imgwidth\textwidth}%
        %\centering
            \includegraphics[width=\textwidth]{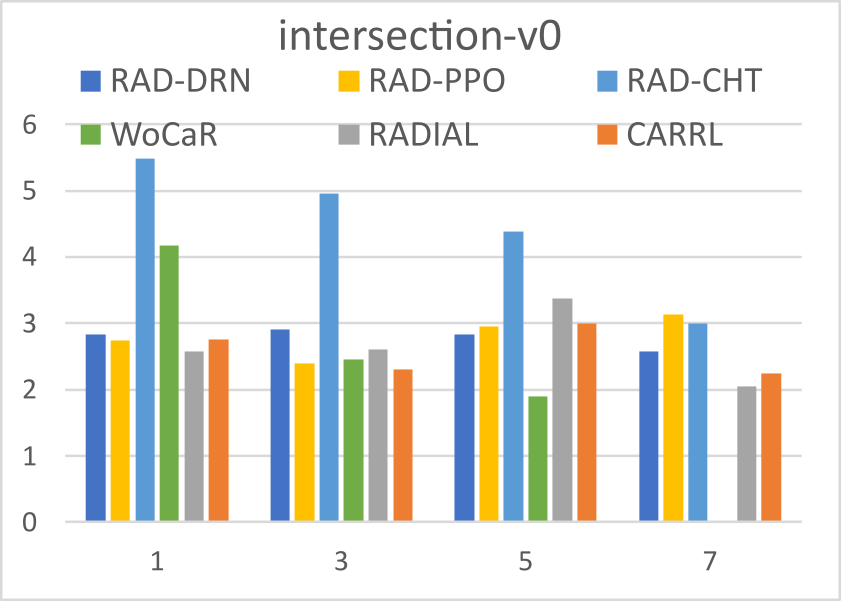}%
            \label{fig:timed:int}
        \end{subfigure}%
        \hspace{0em}\begin{subfigure}[b]{\imgwidth\textwidth}%
        %\centering
            \includegraphics[width=\textwidth]{media/_hopper_timed.png}%
            \label{fig:timed:hopper}
        \end{subfigure}%
        %\hspace{-2.2em}\begin{subfigure}[b]{\imgwidth\textwidth}%
        %\centering
            %\includegraphics[width=\textwidth]{media/halfcheetah_timed.pdf}%
            %\label{fig:timed:halfcheetah}
        %\end{subfigure}%
        \hspace{0em}\begin{subfigure}[b]{\imgwidth\textwidth}%
        %\centering
            \includegraphics[width=\textwidth]{media/_walker_timed.png}%
            \label{fig:timed:walker}
        \end{subfigure}%
        \hspace{0em}\begin{subfigure}[b]{\imgwidth\textwidth}%
        %\centering
            \includegraphics[width=\textwidth]{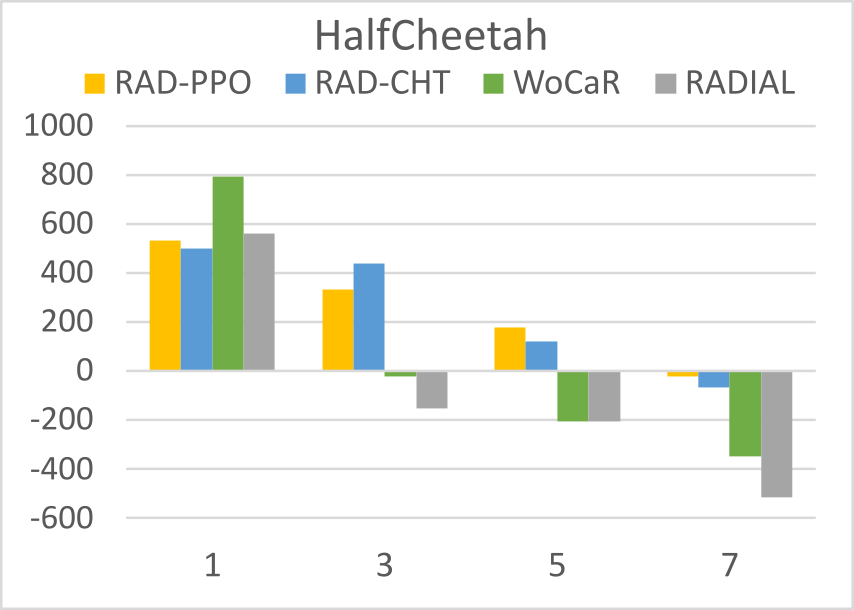}%
            \label{fig:timed:hc}
        \end{subfigure}%
        \hspace{0em}\begin{subfigure}[b]{\imgwidth\textwidth}%
        %\centering
            \includegraphics[width=\textwidth]{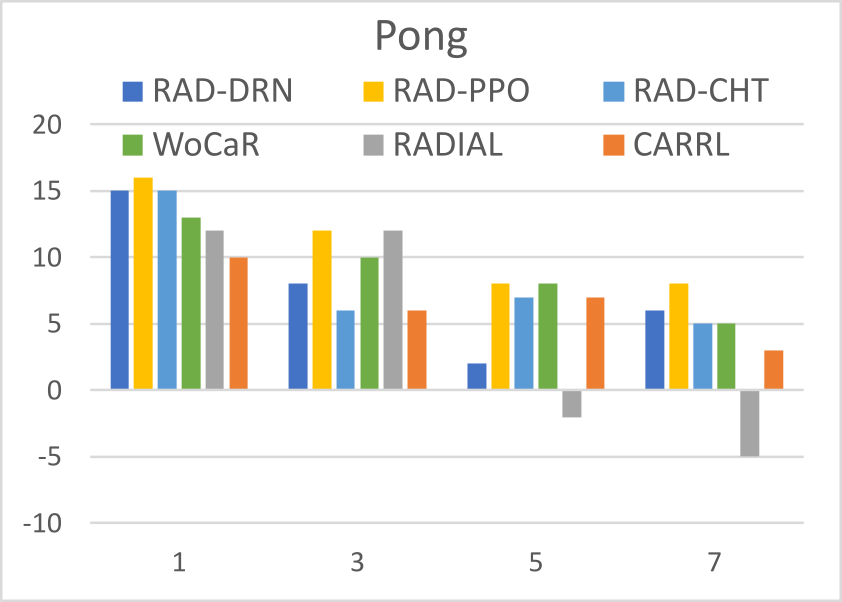}%
            \label{fig:timed:pong}
        \end{subfigure}%
        \hspace{0em}\begin{subfigure}[b]{\imgwidth\textwidth}%
        %\centering
            \includegraphics[width=\textwidth]{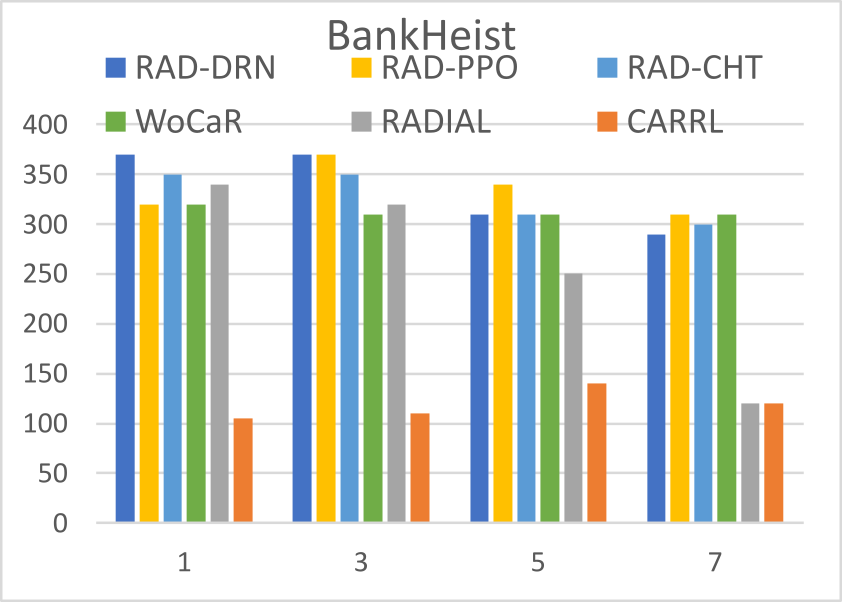}%
            \label{fig:timed:bankheist}
        \end{subfigure}%
        \hspace{0em}\begin{subfigure}[b]{\imgwidth\textwidth}%
        %\centering
            \includegraphics[width=\textwidth]{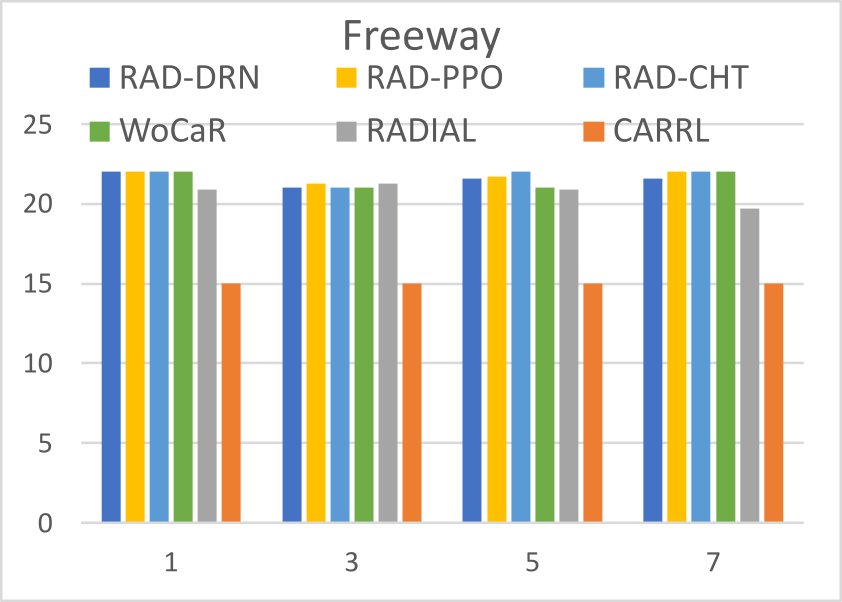}%
            \label{fig:timed:freeway}
        \end{subfigure}%
        \hspace{0em}\begin{subfigure}[b]{\imgwidth\textwidth}%
        %\centering
            \includegraphics[width=\textwidth]{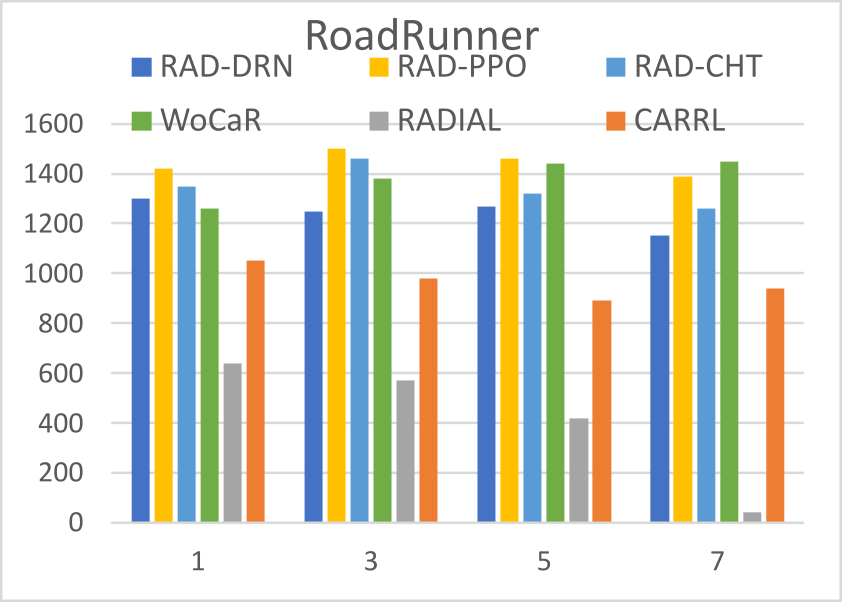}%
            \label{fig:timed:roadrunner}
        \end{subfigure}%
        \caption{Score vs. Percentile Perturbation Frequency with Strategically Timed Attack}%
        \label{fig:timed_attacks}
    \end{subfigure}
    \begin{subfigure}[b]{\textwidth}
        \hspace{0em}\begin{subfigure}[b]{\imgwidth\textwidth}
            \includegraphics[width=\textwidth]{media/_hwy_crit.png}
            \label{fig:crit:hwy}
        \end{subfigure}%
        %\hfill
        \hspace{0em}\begin{subfigure}[b]{\imgwidth\textwidth}
            \includegraphics[width=\textwidth]{media/_mrg_crit.png}
            \label{fig:crit:mrg}
        \end{subfigure}%

        %\hfill
        
        \hspace{0em}\begin{subfigure}[b]{\imgwidth\textwidth}%
        %\centering
            \includegraphics[width=\textwidth]{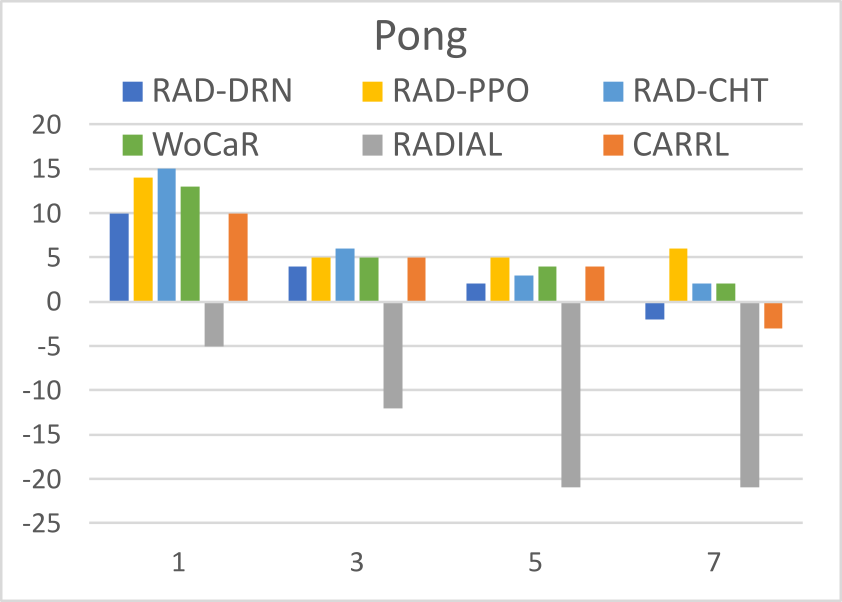}%
            \label{fig:timed:pong}
        \end{subfigure}%
        \hspace{0em}\begin{subfigure}[b]{\imgwidth\textwidth}%
        %\centering
            \includegraphics[width=\textwidth]{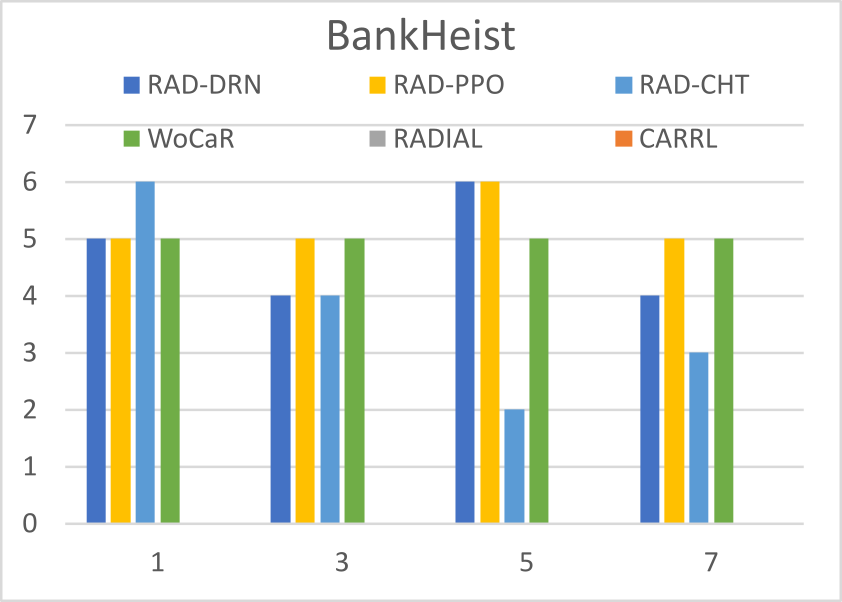}%
            \label{fig:timed:bank}
        \end{subfigure}%
        \hspace{0em}\begin{subfigure}[b]{\imgwidth\textwidth}%
        %\centering
            \includegraphics[width=\textwidth]{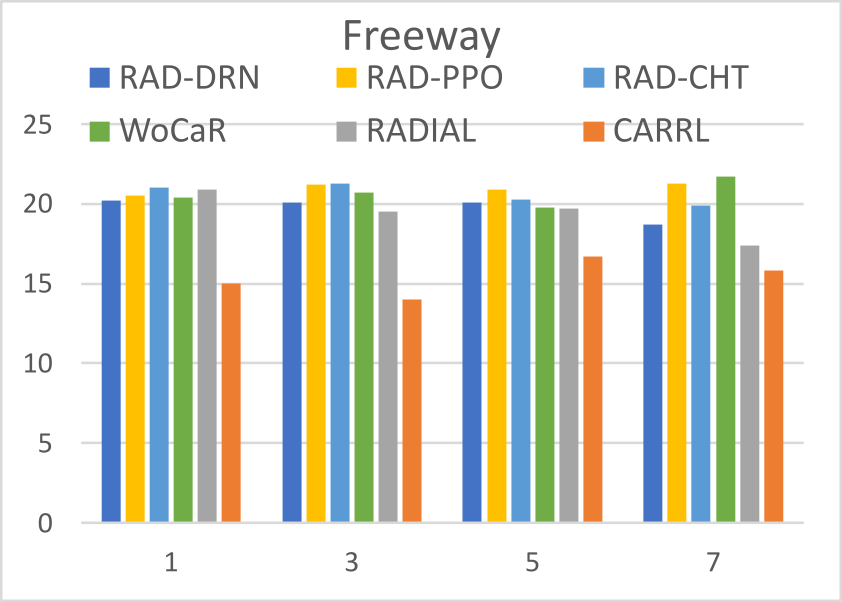}%
            \label{fig:timed:freeway}
        \end{subfigure}%
        \hspace{0em}\begin{subfigure}[b]{\imgwidth\textwidth}%
        %\centering
            \includegraphics[width=\textwidth]{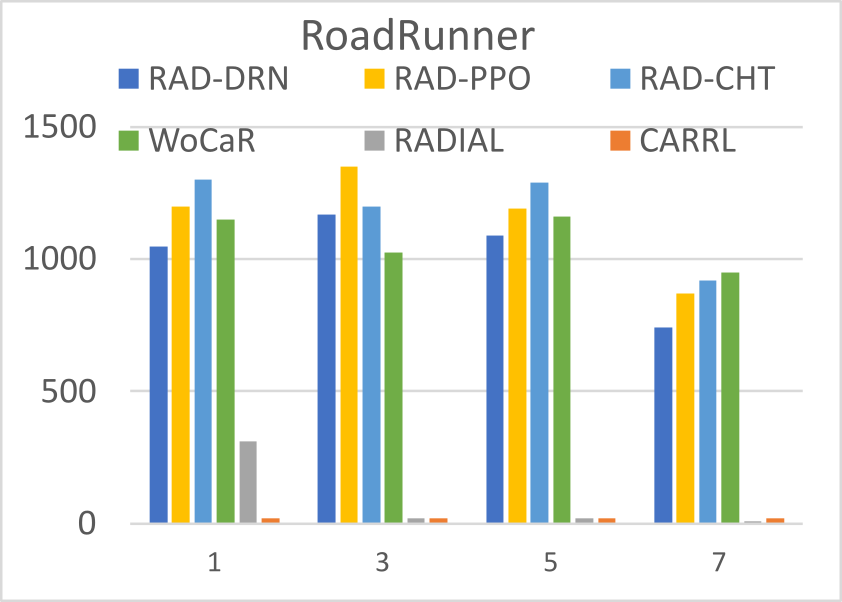}%
            \label{fig:timed:roadrunner}
        \end{subfigure}%
        %\hfill

        \hspace{0em}\begin{subfigure}[b]{\imgwidth\textwidth}
            \includegraphics[width=\textwidth]{media/_hopper_crit.png}
            \label{fig:crit:hopper}
        \end{subfigure}%
        %\hfill
        \hspace{0em}\begin{subfigure}[b]{\imgwidth\textwidth}
            \includegraphics[width=\textwidth]{media/_walker_crit.png}
            \label{fig:crit:walker}
        \end{subfigure}\\%
        \hspace{0em}\begin{subfigure}[b]{\imgwidth\textwidth}%
            \includegraphics[width=\textwidth]{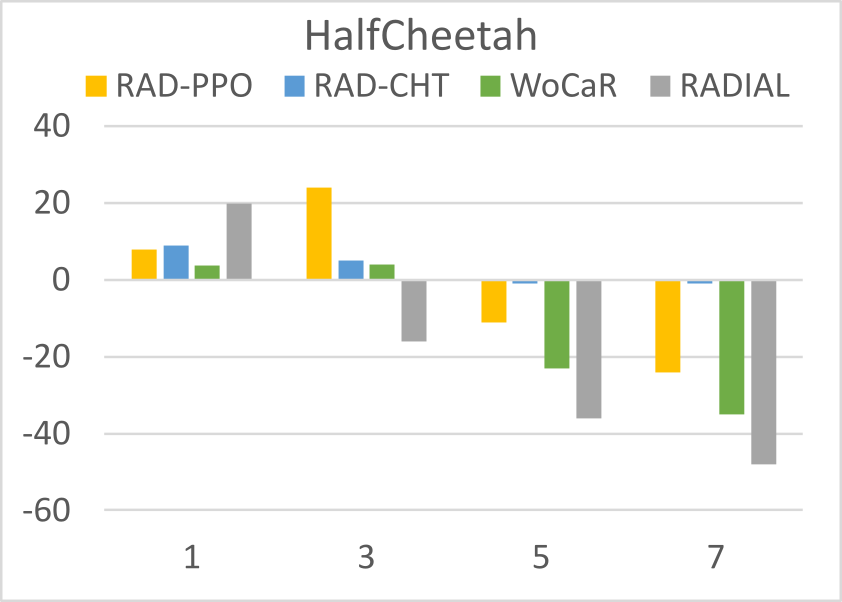}%
            \label{fig:timed:hc}
        \end{subfigure}%
        \hspace{0em}\begin{subfigure}[b]{\imgwidth\textwidth}%
        %\centering
            \includegraphics[width=\textwidth]{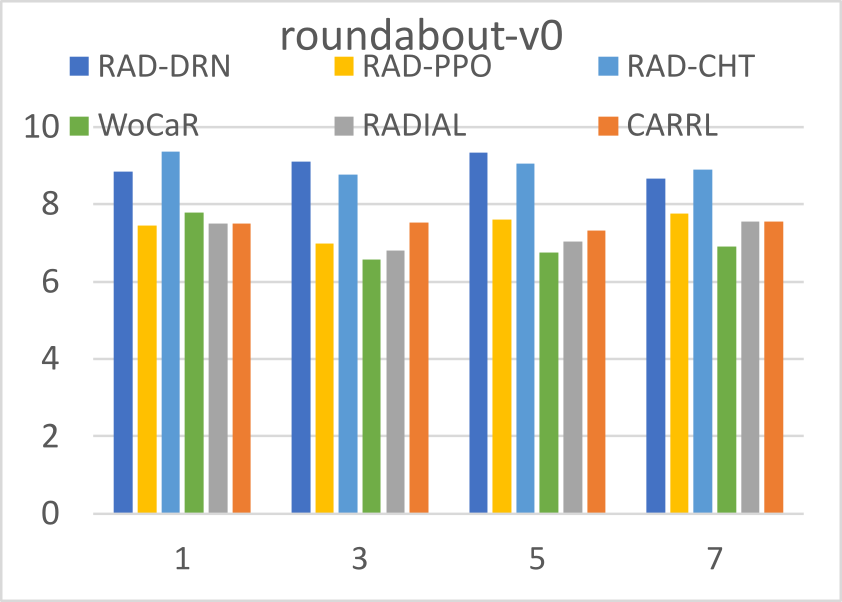}%
            \label{fig:timed:rdb}
        \end{subfigure}%
        \hspace{0em}\begin{subfigure}[b]{\imgwidth\textwidth}%
        %\centering
            \includegraphics[width=\textwidth]{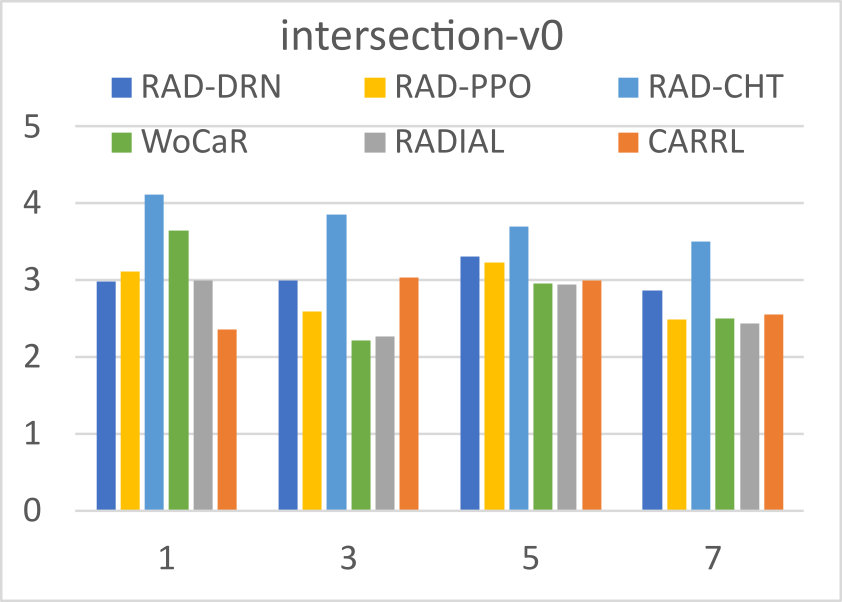}%
            \label{fig:timed:int}
        \end{subfigure}%
        
        \caption{Score vs. Percentile Perturbation Magnitude with Critical Point Attack}%
        \label{fig:timed_attacks}
    \end{subfigure}
    \caption{The performance of robust RL methods against strategic adversaries. The y-axis represents the score and the x-axis represents the intensity of the attack.}
    \label{fig:multistep}
\end{figure*}

\section*{Appendix C: Training Details and Hyperparameters}
\subsection{Model Architecture}
Our RAD-DRN model follows commonly found settings for a simple DQN: a linear input layer, a 64x hidden layer, and a fully connected output layer. For images, we instead use a convolutional layer with an 8x8 kernel, stride of 4 and 32 channels, a convolutional layer with a 4x4 kernel, stride of 2 and 64 channels, and a final convolutional layer with a 3x3 kernel, stride of 1 and 64 channels (this is the same setup as RADIAL and WoCaR implementations). Each layer is followed by a ReLU activation, and finally feeds into a fully connected output. \par
Our RAD-PPO model is built on the standard implementation in StableBaselines3 PPO, with [256,256] MLP network parameters for non-image tasks and [512,512] CNN parameters for Atari domains. To maintain an estimate of the minimum reward to calculate regret, we attach a DQN as constructed above. \par
Worst-case policy attack models are also DQN models as described above, with the output layer size (i.e. number of actions) corresponding to the size of the attack neighborhood.

\subsection{Regarding Neighborhoods}
In our paper, we commonly reference the neighborhood $N(z_t)$ surrounding the observed state $z_t$; This is akin to the $L_p$-ball around $z_t$. In our experiments, we use an $L_p$-ball of radius $\epsilon=3/255$, and sample 10 states to create our neighborhood. We selected the sample size 10 after trying [2,5,10,20] and finding 10 to be sufficient without noticeably impacting computation speed. We also note that changing $\epsilon$ to $2/255$ or $5/255$ has little impact on the final performance of the model.

\subsection{Training Details}
We train our models for 3000 episodes (~10k frames) on highway environments and 1000 episodes (3 million frames) on Atari environments; For MuJoCo simulations, we find 200,000 frames to be sufficient. RAD-DRN learning rates are set to $5e-4$ for all environments (tuned from a range $[1e-3, 3e-4, 5e-4, 5e-5, 5e-7]$, while RAD-PPO uses the default parameters of StableBaselines3's PPO implementation. Worst-case adversaries are trained very briefly as they are solving a rather simple MDP, we use $10,000$ frames for both highway and Atari environments and use a learning rate of $0.05$. For release models, we use the same training settings as WocaR: $1,000$ training episodes, resulting in around $2,000,000$ training frames. Default settings are available in our implementation as well as the preceding work.
\subsection{Testing Details}
During the testing of our models, we repeat our experiments 10 times each for 50 random seeds, for a total of 500 episodes in a test. \par
During testing, adversary agents are permitted to make perturbations every $t_{adv}$ steps, to reduce the rate of critical failure (which is unhelpful to the evaluation of the methods). Our experiments on highway environments are conducted with $t_{adv} = 2$, and $t_{adv}=5$ on Atari domains. To remove temporal dependence, this interval is shifted by one between testing episodes---the victim is tested against perturbations at every step at least once, but not every step in a single episode.  \par

%%%%%%%%%%%%%%%%%%%%%%%%%%%%%%%%%%%%%%%%%%%%%%%%%%%%%%%%%%%%%%%%%%%%%%%%

\end{document}